\documentclass[lettersize,journal]{IEEEtran}
\usepackage{graphicx}
\usepackage{cite}
\usepackage{picinpar}
\usepackage{amsmath}
\usepackage{url}
\usepackage{xcolor}
\usepackage{flushend}
\usepackage[latin1]{inputenc}
\usepackage{colortbl}
\usepackage{soul}
\usepackage{multirow}
\usepackage{pifont}
\usepackage{color}
\usepackage{alltt}
\usepackage[hidelinks]{hyperref}
\usepackage{enumerate}
\usepackage{siunitx}
\usepackage{breakurl}
\usepackage{epstopdf}
\usepackage{pbox}
\usepackage{amssymb}
\usepackage{subfigure}
\usepackage{booktabs}
\usepackage{threeparttable}
\usepackage{algorithm}
\usepackage{algpseudocode}
\usepackage{caption}
\usepackage{amsthm}
\theoremstyle{plain}
\newtheorem{theorem}{Theorem}[section]

\theoremstyle{definition}

\newtheorem{assumption}[theorem]{Assumption}
\theoremstyle{remark}

\begin{document}
\title{CROP: Conservative Reward for Model-based Offline Policy Optimization}

\author{
	\vskip 1em
	
	Hao Li,
	Xiao-Hu Zhou, \emph{Member, IEEE},
        Shu-Hai Li,
	Mei-Jiang Gui,
	Xiao-Liang Xie,\emph{Member, IEEE},
	Shi-Qi Liu, \\
	Shuang-Yi Wang, Zhen-Qiu Feng, and Zeng-Guang Hou, \emph{Fellow, IEEE}

	\thanks{
	
        This work was supported in part by the National Key Research and Development Program of China under Grant 2023YFC2415100, in part by the National Natural Science Foundation of China under Grant 62373351, Grant 62503474, Grant 82327801, Grant 62303463, in part by the Chinese Academy of Sciences Project for Young Scientists in Basic Research under Grant No. YSBR-104, in part by the Beijing Natural Science Foundation under Grant F252068, Grant 4254107, in part by Beijing Nova Program under Grant 20250484813, in part by China Postdoctoral Science Foundation under Grant 2024M763535, in part by the Postdoctoral Fellowship Program of CPSF under Grant GZC20251170, and in part by Xiaomi Young Talents Program/Xiaomi Foundation. (Corresponding authors: Xiao-Hu Zhou and Zeng-Guang Hou)
		
		H. Li, X.-H. Zhou, S.-H. Li, M.-J. Gui, X.-L. Xie, S.-Q. Liu, S.-Y. Wang, Z.-Q. Feng, and Z.-G. Hou are with the State Key Laboratory of Multimodal Artificial Intelligence Systems, Institute of Automation, Chinese Academy of Sciences, Beijing 100190, China (e-mail: lihao2020@ia.ac.cn; xiaohu.zhou@ia.ac.cn; zengguang.hou@ia.ac.cn).

        \copyright 2026 IEEE.  Personal use of this material is permitted.  Permission from IEEE must be obtained for all other uses, in any current or future media, including reprinting/republishing this material for advertising or promotional purposes, creating new collective works, for resale or redistribution to servers or lists, or reuse of any copyrighted component of this work in other works.
	}
}

\maketitle


\definecolor{limegreen}{rgb}{0.2, 0.8, 0.2}
\definecolor{forestgreen}{rgb}{0.13, 0.55, 0.13}
\definecolor{greenhtml}{rgb}{0.0, 0.5, 0.0}
\begin{abstract}
Offline reinforcement learning (RL) aims to optimize a policy using collected data without online interactions. Model-based approaches are particularly appealing for addressing offline RL challenges because of their capability to mitigate the limitations of data coverage through data generation using models. Nonetheless, a prevalent issue in offline RL is the overestimation caused by distribution shift. This study proposes a novel model-based offline RL algorithm named \underline{C}onservative \underline{R}eward for model-based \underline{O}ffline \underline{P}olicy optimization (CROP). CROP introduces a streamlined objective that concurrently minimizes estimation error and the rewards of random actions, thereby yielding a robustly conservative reward estimator.
Theoretical analysis shows that the designed conservative reward mechanism leads to a conservative policy evaluation and mitigates distribution shift.
Experiments showcase that with the simple modification to reward estimation, CROP can conservatively estimate the reward and achieve competitive performance with existing methods. The source code will be available after acceptance.
\end{abstract}

\section{Introduction}
\label{section: introduction}
Reinforcement learning (RL) has achieved impressive performance in various decision-making domains, including electronic games~\cite{DQN}, robot control~\cite{DSAC-AE}, and recommender systems~\cite{Zhang2017DeepLB}. Conventional RL uses an online training paradigm in which the agent optimizes policies based on real-time interactions with the environment~\cite{Sutton2005ReinforcementLA}. However, online interactions can be expensive, time-consuming, or dangerous, posing a significant hurdle for widespread RL applications. To address this issue, a natural idea is to use pre-collected data instead of online interactions in RL, which is known as offline RL~\cite{lange2012batch, offlineRLReview}. 
In practice, offline datasets often suffer from low quality and limited coverage. When standard online RL algorithms are applied directly to such data, they typically yield suboptimal performance - primarily due to distribution shift issues~\cite{CQL}. The distribution shift arises from the discrepancy between the behavior policy used for data collection and the evolving learned policy, causing erroneous overestimation of the Q-function and damaging the performance.

To alleviate the distribution shift, many model-free offline RL algorithms incorporate conservatism or regularization to constrain the learned policy~\cite{CQL, ATAC, BEAR, DBLP:conf/icml/KostrikovFTN21, DBLP:conf/icml/Shi0W0C22, casog}. However, model-free algorithms can only directly learn about the states in the offline data and cannot generalize environmental information, leading to myopia and poor performance in unseen states.

\begin{figure}[t]
\centering
\includegraphics[width=\columnwidth]{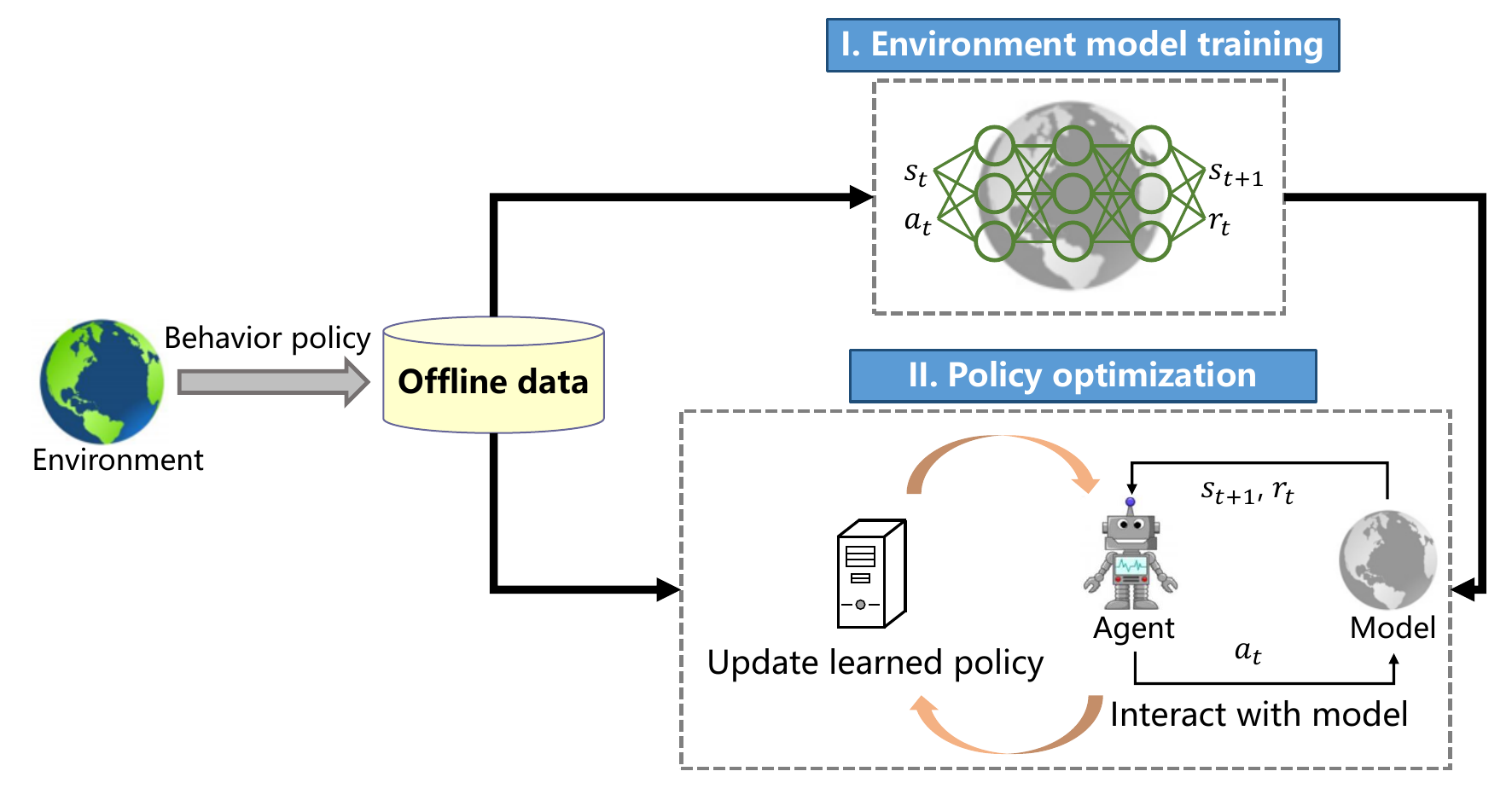} 
\caption{The common paradigm for model-based offline RL.}
\label{fig: framework}
\end{figure}

Model-based offline RL algorithms solve the limitation by training an environment model for interactions with the agent, as shown in Fig.~\ref{fig: framework}. Nevertheless, distribution shift can negatively impact model-based offline RL methods. The model accuracy inherently diminishes for state-action pairs that lie beyond the scope of the offline dataset. These inaccurate situations may be accessed by the agent due to the distribution shift, influencing the policy optimization effect. Several methods estimate model uncertainty and penalize the situations with low model accuracy. However, these methods rely on some sort of strong heuristic assumption about uncertainty estimation~\cite{MOPO, MOBILE, DBLP:conf/iclr/LuBPOR22, DBLP:conf/l4dc/RafailovYRF21, Depeweg2016LearningAP} or directly detect out-of-distribution (OOD) state-action tuples~\cite{MOReL}, which might prove fragile or impractical in complex environments. Furthermore, some researchers have heuristically designed elaborate structures, such as introducing counters~\cite{count} or inverse dynamic functions~\cite{CABI}, to punish OOD data. To eschew uncertainty estimation or other additional parts, several methods induce conservatism into the model~\cite{RAMBO, ARMOR} or Q-function~\cite{COMBO} in policy optimization and underestimate OOD state-action tuples, indirectly mitigating distribution shift.

To address these limitations, we propose a \underline{C}onservative
\underline{R}eward for model-based \underline{O}ffline \underline{P}olicy optimization (CROP) RL algorithm, whose core idea is to directly incorporate conservatism into the reward function rather than the policy or value function. Our main contributions include: 
\begin{enumerate}
    \item[\text{(1)}]
    The proposed method introduces a novel conservative reward estimation that minimizes rewards of random actions alongside the estimation error in model training. The design avoids not only model uncertainty estimators or additional components but also adversarial updates to the environment model during policy optimization.
    \item[\text{(2)}]
    Theoretical analysis demonstrates that this proposed method can underestimate Q-function and mitigate distribution shift. Its stability and performance lower bound analysis are also provided.
    \item[\text{(3)}]
    Experimental evaluations verify the performance of CROP and demonstrate its competitive results among representative offline RL approaches.
\end{enumerate}

The remainder of this paper is organized as follows. Section~\ref{section: related work} reviews related work in offline RL. Section~\ref{section: preliminaries} introduces necessary preliminaries. Section~\ref{section: method} presents the proposed method, including model training, policy optimization, and theoretical analysis. Section~\ref{section: experiment} reports experimental results and ablation studies. Section~\ref{section: discussion} provides discussions on its implications and limitations. Finally, Section~\ref{section: conclusion} concludes the paper and outlines future research directions.

\section{Related Work}
\label{section: related work}

Offline RL presents no opportunity for exploration during policy optimization. When offline data sufficiently cover the state-action space, existing online RL methods can perform effectively without additional modification~\cite{Agarwal2019AnOP}. However, in more common cases, offline data are insufficiently covered.
Using out-of-distribution (OOD) actions is necessary to find better policies, but this approach also brings potential bias and risks, which should be balanced in policy optimization~\cite{offlineRLReview}.
For a single update step, the probability of the OOD value function being underestimated and overestimated is equal. However, since the policy optimization prefers actions with higher value functions and the Bellman equation used in the value function estimation includes bootstrapping, overestimation can accumulate catastrophically~\cite{CQL}.
Online RL methods often perform extremely poorly in such settings~\cite{d4rl, CQL}. In the following section, we discuss how existing offline methods address the overestimation challenge.

\textbf{Model-free offline RL:}
Policy constraint methods introduce hard constraints to either (1) directly constrain the learned policy to stay near the behavior policy~\cite{BCQ, TD3+BC, EMaQ} or (2) prevent OOD actions from being used in the Bellman backup updates~\cite{BEAR, MCQ, IQL}.
These methods directly limit the scope of policy optimization and may perform poorly when the behavioral policy is inferior. By contrast, underestimating the value function can avoid OOD actions indirectly without hard constraint. Such underestimation can be achieved directly by penalizing uncertainty based on Q function ensembles~\cite{EDAC, PBRL} or implicitly by redesigning the loss function~\cite{CQL, ATAC}.

\textbf{Model-based offline RL:} Model-based offline RL methods initially train an environment model using offline data. During policy optimization, the learned policy interacts with the trained models to extend the offline data. Such a training design enables the agent to access states that do not occur in offline data~\cite{MOPO}. Policy constraint methods also have a role in model-based settings~\cite{MOOSE}.
Moreover, model uncertainty is commonly used as a penalty term in offline RL~\cite{MOPO, MOBILE, DBLP:conf/iclr/LuBPOR22, DBLP:conf/l4dc/RafailovYRF21, Depeweg2016LearningAP, DBLP:journals/corr/abs-2312-03991}. However, such methods often necessitate strong prior assumptions during uncertainty quantification.
For example, MOPO~\cite{MOPO} assumes the predicted variance as a reasonable estimation of model uncertainty. These assumptions may be unreliable for neural network models~\cite{Gawlikowski2021ASO, DBLP:conf/iclr/LuBPOR22}.
In addition, recent research has designed additional components, such as discriminators~\cite{MOReL}, inverse models~\cite{CABI}, and counters~\cite{count}, and incorporated them into the model to conservatively estimate OOD action.
However, additional components complicate the model and compromise the speed and stability.
To avoid model uncertainty estimators or additional components, some studies modify the loss function to introduce conservatism into Q-function or the environment model~\cite{COMBO, RAMBO, ARMOR}.
Some methods use robust RL to improve the performance in the worst cases among all possibilities, and design adversarial structures to train the environment model to estimate the worst cases~\cite{RAMBO, ARMOR}.

The proposed method CROP modifies the loss function to implicitly introduce conservatism as several existing model-based offline RL methods (COMBO~\cite{COMBO}, RAMBO~\cite{RAMBO}, and ARMOR~\cite{ARMOR}).
COMBO is a model-based extension of CQL~\cite{CQL}, while RAMBO and ARMOR combine the robust RL with offline RL.
CROP differs from existing methods in three main aspects:
\begin{itemize}
    \item
    The existing three methodologies are specifically engineered to directly underestimate the Q function. In contrast, CROP focuses on the conservative estimation of rewards.
    \item 
    While the existing methods implicitly introduce conservatism in the Q-function (COMBO) or the entire environment model (RAMBO and ARMOR), CROP only introduces conservatism in the reward estimator of the environment model.
    \item
    CROP only revises the loss in model training to obtain a conservative reward estimator, whereas these methods redesign the loss in policy optimization. Compared with adversarial robust RL in policy optimization~\cite{RAMBO}, this design is simpler and more time-saving.
\end{itemize}

\section{Preliminaries}
\label{section: preliminaries}

RL is used for optimization problems in Markov Decision Process (MDP).
An MDP is defined by a tuple $\left(\mathcal{S}, \mathcal{A}, T, R, \mu_0, \gamma \right)$, where $\mathcal{S}$ and $\mathcal{A}$ represent the state and action spaces, respectively. $T$, $R$, $\mu_0$, and $\gamma$ denote the transition probability, reward, initial state distribution, and discount factor, respectively. At time step $t$ with state $s\in \mathcal{S}$, the agent selects an action $a$ based on a policy $\pi(a|s)$. Then, the state changes from $s$ to $s'$ based on the transition probability $T(s'|s, a)$, and the agent obtains a reward $R(s, a)\in \left[ -R_{\max}, R_{\max}\right]$. The goal of RL is to find the optimal policy that maximizes the expected cumulative discounted reward $\sum_t \mathbb{E}_{s,a}\left[\gamma^t R(s, a)\right]$ in the MDP. 
Offline RL is a special formulation of RL that only uses a previously collected dataset $\mathcal{D} = \left\{(s, a, R(s,a), s')\right\}$ during training~\cite{offlineRLReview}. We use $\bar{\pi}$ to denote the empirical behavior policy in $\mathcal{D}$.
\section{Method}
\label{section: method}
\begin{algorithm}[t!]
    \caption{\underline{C}onservative \underline{R}eward for model-based \underline{O}ffline \underline{P}olicy optimization (CROP)}
	\label{alg: CROP} 
	\renewcommand{\algorithmicrequire}{\textbf{Input:}}
	\renewcommand{\algorithmicensure}{\textbf{Output:}}
	\begin{algorithmic}[1]
	    \State {\bfseries Input:} offline data $\mathcal{D} = \left\{(s, a, R(s,a), s')\right\}$
	    \Repeat 
	    \State divide $\mathcal{D}$ into a training set and a validation set;
		\Repeat
		\State train $\hat{T}$ using Equation~(\ref{equation: transition loss});
		\Until the loss no longer falls on the validation set
		\Repeat
		\State train $\hat{r}$ using Equation~(\ref{equation: reward loss});
		\Until the loss no longer falls on the validation set
		\Until
		\State recompute the reward in $\mathcal{D}$;
		\Repeat 
		\State interact with the model using policy $\pi$;
		\State update Q-function $\hat{Q}^{\pi}$ using Equation~(\ref{equation: q loss});
		\State update policy $\pi$ using Equation~(\ref{equation: pi loss});
		\State update $\alpha$ using Equation~(\ref{equation: alpha loss});
		\State update the target Q-function $Q_{\mathrm{tar}}$;
		\Until{}
	\end{algorithmic}
	
\end{algorithm}

\subsection{Model training with conservative reward estimation}
\label{subsection: model training}

\begin{figure}[t]
\centering
\includegraphics[width=\columnwidth]{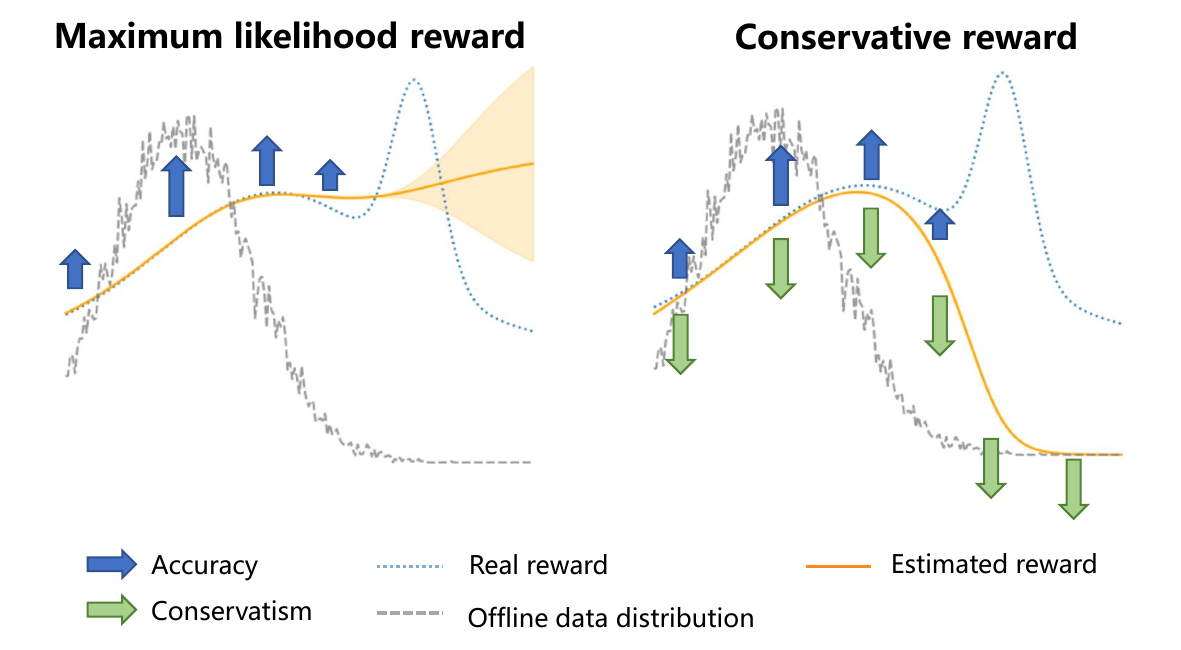} 
\caption{Because the data within the distribution needs to trend towards the real reward, the proposed conservative reward based on minimizing the reward of random actions can more obviously underestimate OOD actions.}
\label{fig: idea}
\end{figure}

The core idea of CROP is to deliberately underestimate rewards for OOD actions to mitigate distribution shift. This is implemented by jointly minimizing both the estimation error and rewards of random actions when updating the reward estimator $\hat{r}$:
\begin{equation}
    \label{equation: reward loss}
    l_{r} = \mathbb{E}_{\mathcal{D}}\left\{\left[ \hat{r}(s,a) - R(s, a)\right]^2 + \beta*{\rm mean} \left[\hat{r}(s,\bar{a})\right]\right\}.
\end{equation}
$\bar{a}$ denotes random actions, and hyperparameter $\beta$ controls the level of conservatism.
Intuitively, the former and latter correspond to the correctness and conservatism of reward estimations, respectively. OOD actions are almost unaffected by correctness and are more significantly underestimated as in Fig.~\ref{fig: idea}.
By setting the derivative of Equation~(\ref{equation: reward loss}) to zero, we obtain the optimal conservative reward estimation $r$:
\begin{equation}
    \label{equation: reward}
        r\left(s,a\right) = R\left(s,a\right) - \beta \frac{\mu}{\bar{\pi}\left(a|s\right)},
\end{equation}
where $\mu$ is the probability density of the uniform distribution in action space $\mathcal{A}$, and $\bar{\pi}$ is the behavior policy of $\mathcal{D}$. The second term on RHS of Equation~(\ref{equation: reward}) represents the conservativeness of the reward estimation and is inversely proportional to the probability of the action appearing in $\mathcal{D}$, indicating that OOD actions are underestimated more strongly. 
Although a similar form of conservative reward has been derived in the context of robust MDPs \cite{Policygradientforrectangular}, our application and motivation are specific to robust RL. The proposed method punishes rewards of random actions to underestimate the unknown situations, while robust RL uses adversarial learning to consider the worst case.

CROP does not modify the update of the state transition estimator and policy. The transition probability estimator $\hat{T}$ is updated by maximizing log-probability, which is a classic design in model-based RL:
\begin{equation}
    \label{equation: transition loss}
    l_{T} = \mathbb{E}_{\mathcal{D}}\left\{ -\log \hat{T}(s'|s, a)\right\}.
\end{equation}
By interacting with models using the conservative reward estimation $\hat{r}$, conventional RL algorithms can avoid OOD actions and improve policies safely in offline settings, which will be analyzed theoretically in Section~\ref{subsection: theoretical analysis}. Therefore, CROP provides a new perspective to connect offline and online RL whereby offline RL can be regarded as online RL under conservative reward estimation, which may help to apply the appealing development of online RL to offline RL problems.

\subsection{Practical implementation}
\label{subsection: practical implementation}
We describe a practical implementation of CROP using the conservative reward estimation described in Section~\ref{subsection: model training}.
The algorithm is summarized in Algorithm~\ref{alg: CROP}, which consists of model training (line 2 to 10) and policy optimization (line 12 to 18).

In model training, we learn an ensemble of models, and each model is trained independently. For each model, the offline data $\mathcal{D}$ are divided into a train set and a validation set individually, and then $\hat{T}$ and $\hat{r}$ are trained using Equation~(\ref{equation: transition loss}) and Equation~(\ref{equation: reward loss}), respectively. ${\rm mean} \left[\hat{r}(s,\bar{a})\right]$ is calculated using $n$ random actions. In practice, we find that $\hat{r}$ of some actions tends to approach negative infinity, which greatly affects the effectiveness and stability. This occurs because offline data cannot cover all actions in the continuous action space. Some actions satisfy $\bar{\pi}(a|s) = 0$, causing $\bar{r}$ to approach negative infinity. To solve this problem, the output of the reward estimator is mapped to $[0,1]$ via sigmoid and then linearly transformed to $[r_{\min}, r_{\max}]$, where $r_{\min}$ and $r_{\max}$ are the minimum and maximum values of the reward in the offline data, respectively.

After model training, the reward in offline data is replaced by the mean of the reward predictor ensemble $\hat{r}$.
Then, an online model-free RL algorithm, Soft Actor-Critic (SAC)~\cite{SAC}, is used to optimize the policy from offline data and online interactions with the model ensemble. In interactions with the model ensemble, the reward is computed as the mean of $\hat{r}$, whereas the next state is sampled from the output of $\hat{T}$ in a model selected at random.
Each interaction episode begins with an initial state sampled uniformly from the offline dataset $\mathcal{D}$ and proceeds for a fixed horizon of $k$ steps.
During each policy optimization step, a mini-batch data $\mathcal{D}_{f}$ is sampled, where the proportion of model-generated online interactions is $f$.
Q-function of policy $\pi$, which is denoted by $\hat{Q}^{\pi}$, is trained by minimizing the soft Bellman residual:
\begin{equation}
    \label{equation: q loss}
    l_{\hat{Q}^{\pi}} = \mathbb{E}_{\mathcal{D}_f} \left\{ \left[\hat{Q}^{\pi}\left(s, a\right) - \hat{r}\left(s, a\right) - \gamma \hat{V}^{\pi}\left(s'\right)\right]^2  \right\}.
\end{equation}
$\hat{V}^{\pi}$ is estimated by Monte Carlo method:
\begin{equation}
    \label{equation: v}
    \hat{V}^{\pi}(s) = \mathbb{E}_{a \sim \pi(\cdot|s)} \left[Q_{\mathrm{tar}}\left(s, a\right) - \alpha  \log \pi(a|s) \right],
\end{equation}
where the parameters of target Q-function $Q_{\mathrm{tar}}$ are an exponentially moving average of Q-function parameters.
The policy $\pi$ is trained by maximizing the value function $V^{\pi}$ and maintaining a reasonable entropy with a Lagrangian relaxation.
The policy loss function $l_{\pi}$ is as follows:
\begin{equation}
    \label{equation: pi loss}
    l_{\pi} = \mathbb{E}_{\mathcal{D}_f}\left\{\alpha \mathcal{H}\left[\pi\left(\cdot|s\right)\right]-\hat{V}^{\pi}\left(s\right)\right\}, 
\end{equation}
where $\mathcal{H}(\cdot)$ stands for entropy.
$\alpha$ is a non-negative parameter updated by minimizing the following loss function:
\begin{equation}
    \label{equation: alpha loss}
    l_{\alpha} = \mathbb{E}_{\mathcal{D}_f}\left\{\alpha\mathcal{H}\left[\pi\left(\cdot|s\right)\right]-\alpha\bar{\mathcal{H}}\right\},
\end{equation}
where hyperparameter $\bar{\mathcal{H}}$ is the target entropy.

$\hat{T}$, $\hat{r}$, $Q_{\rm{tar}}$, $\hat{Q}^{\pi}$, and $\pi$ are parameterized with multi-layer perceptrons.

Although the conservative reward estimation suppresses Q-value overestimation for out-of-distribution actions, it may potentially limit the diversity of learned policies. To address this, CROP includes multiple components to balance exploration and exploitation:
\begin{enumerate}
    \item[\text{$\bullet$}]
    The hyperparameter $\beta$ controls the degree of conservatism. A small $\beta$ encourages broader action space coverage, while a large $\beta$ prioritizes safety.
    \item[\text{$\bullet$}]
    The rollout buffer is composed of a mixture of real data and model-generated data, controlled by the proportion of model
    generated online interactions $f$. This hybrid composition allows for controlled exploration in the model-based environment without deviating too far from known distributions.
    \item[\text{$\bullet$}]
    The policy is trained using SAC, which incorporates an entropy term that encourages policy diversity and avoids premature convergence.
\end{enumerate}
These components collectively enable CROP to maintain policy safety while allowing for adequate improvement over suboptimal behavior policies.

One potential concern in model-based offline RL is that model-generated data may introduce bias due to prediction errors, thus worsening the distribution shift. To mitigate this, CROP integrates several safeguards: (1) model ensembles are used to reduce variance, (2) early stopping on a validation set avoids overfitting, (3) reward predictions are clipped via sigmoid transformation, and (4) short rollout lengths are used to reduce error accumulation. Most importantly, the conservative reward estimation penalizes out-of-distribution actions more heavily, thereby discouraging the policy from relying on unreliable generated samples. This design acts as an implicit filter, mitigating the risk of compounding model errors during policy training.

\subsection{Theoretical analysis of CROP}
\label{subsection: theoretical analysis}

In the following,  we theoretically analyze the proposed method, CROP, and show that it underestimates Q-function and has a performance lower bound. Although we discuss the tabular case for clarity and tractability, the core insights extend naturally to continuous state and action spaces. The conservative reward estimation framework remains applicable by replacing probability terms with density functions and summations with integrals. For more explanations, see Appendix~\ref{appendix:A}.

Let $\bar{T}$ and $\bar{r}$ denote the empirical transition probability and the empirical conservative reward estimated from the dataset $\mathcal{D}$. These represent the optimal achievable approximations of the true $T$ and $r$ from the finite offline data. The difference between $(\bar{T}, \bar{r})$ and $(T, r)$ comes from the sample bias during offline data collection, whereas the difference between $(\bar{T}, \bar{r})$ and $(\hat{T}, \hat{r})$ comes from the estimation error of models (neural networks in this paper). 
The sample bias and the estimation error are the main factors affecting the performance of the algorithm.
To express the Q-function update conveniently, we use $B^{T,r,\pi}$ to denote the Bellman operator about policy $\pi$ in the MDP with transition probability $T$ and reward $r$:
\begin{equation}
    \label{equation: bellman}
    B^{T,r,\pi}Q^{\pi} = r +\gamma T^{\pi} Q^{\pi},
\end{equation}
where $T^{\pi}$ denotes the state transition probability with policy $\pi$: $T^{\pi}(s'|s)=\mathbb{E}_{a \sim \pi(\cdot|s)} [T(s'|s,a)]$. 

Following the standard assumption in model-based offline RL literature~\cite{COMBO, SafePolicyImprovement}, we assume that the sample bias and the estimation error are bounded as follows:
\begin{assumption}
\label{assumption: 1}
$\forall s,a\in D$, the following relationships hold with high probability, $\geq 1-\delta$
\begin{equation}
\begin{split}
    |\bar{r}\left(s,a\right)-r\left(s,a\right)| \leq \frac{C_{r,\delta}}{\sqrt{|D\left(s,a\right)|}},\\ ||\bar{T}\left(s'|s,a\right)-T\left(s'|s,a\right)||_1\leq \frac{C_{T,\delta}}{\sqrt{|D\left(s,a\right)|}},
\end{split}
\end{equation}
\end{assumption}

\begin{assumption}
\label{assumption: 2}
$\forall s,a\in D$, the model estimation bias is bounded:
\begin{equation}
    |\hat{r}\left(s,a\right)-\bar{r}\left(s,a\right)| \leq \epsilon_{r}, ||\hat{T}\left(s'|s,a\right)-\bar{T}\left(s'|s,a\right)||_1\leq \epsilon_{T}
\end{equation}
\end{assumption}

The estimated Q-function $\hat{Q}^{\pi}$ is computed on $D_f$; thus, $\hat{Q}^{\pi}$ is the fixed point of $B^{T_f,\hat{r},\pi}$, where $T_f = (1-f) \bar{T} + f\hat{T}$. 
First, CROP is stable during calculation:
{
\setcounter{theorem}{0}
\begin{theorem}
\label{proposition: 0}
Let $\|\cdot\|_{\infty}$ be the $L_{\infty}$ norm and \\ $(R^{|S\times A|},\|\cdot\|_{\infty})$ be a complete space. Then, Bellman operator of CROP $B_{\text{CROP}} = B^{T_f,\hat{r},\pi}$ is a $\gamma$-contraction mapping operator; i.e., 
\begin{equation*}
    \|B_{\text{CROP}}Q_{1} - B_{\text{CROP}}Q_{2}\|_{\infty} \leq \gamma \|Q_{1} - Q_{2}\|_{\infty}.
\end{equation*}
\end{theorem}
\begin{proof}
We aim to prove that the Bellman operator used in the CROP method, 
\[
B_{\text{CROP}} Q(\ast) = \left[(1 - f)B^{\bar{T}, \hat{r}, \pi} + fB^{\hat{T}, \hat{r}, \pi} \right] Q(\ast),
\]
is a contraction mapping, where $*$ denotes $(s,a)$. Recall the general form of a policy evaluation Bellman operator, $B^{T, r, \pi} Q = r + \gamma \, T^{\pi} Q$. Since \(T^{\pi}\) performs a weighted average on \(Q\), it is a non-expansive mapping, that is, $\left\|T^{\pi}Q_{1}-T^{\pi}Q_{2}\right\|_{\infty} \leq\left\|Q_{1}-Q_{2}\right\|_{\infty}$. Therefore, both $B^{\bar{T}, \hat{r}, \pi}$ and $B^{\hat{T}, \hat{r}, \pi}$ are $\gamma$-contraction mappings. 

Define the CROP operator as a convex combination of the two contractions, with $f \in [0,1]$. Then for any $Q_1$, $Q_2$:
\begin{equation}
\begin{aligned}
    \| B_{\text{CROP}} Q_1 - B_{\text{CROP}} Q_2 \|_\infty \\
    =\left\| (1-f)(B^{\bar{T}, \hat{r}, \pi} Q_1 - B^{\bar{T}, \hat{r}, \pi} Q_2)\right. \\
    + \left. f(B^{\hat{T}, \hat{r}, \pi} Q_1 - B^{\hat{T}, \hat{r}, \pi} Q_2) \right\|_\infty\\
     \leq (1-f)\gamma \| Q_1 - Q_2 \|_\infty + f\gamma \| Q_1 - Q_2 \|_\infty \\
    = \gamma \| Q_1 - Q_2 \|_\infty,
\end{aligned}
\end{equation}

Thus, $B_{\text{CROP}}$ is also a $\gamma$-contraction. According to the fixed point theorem, any initial Q function can converge to a unique fixed point by repeatedly applying the conservative Bellman operator $B_{\text{CROP}}$ .
\end{proof}
}

We state that $\hat{Q}^{\pi}$ conservatively estimates the true Q-function:
\begin{theorem}
\label{theorem: 1}
For sufficiently large $\beta$, we have
\begin{equation}
    \mathbb{E}_{s,a} \left[\hat{Q}^{\pi}\left(s,a\right)\right]\leq\mathbb{E}_{s,a} \left[Q^{\pi}\left(s,a\right)\right],
\end{equation}
where $Q^{\pi}$ is the Q-function of $\pi$ in the actual MDP, i.e., the fixed point of $B^{T, R,\pi}$.
\end{theorem}
\begin{proof}

With Assumption~\ref{assumption: 1}, the difference between $B^{\bar{T},\bar{r},\pi}$ and $B^{T,r,\pi}$ can be bounded:
\begin{equation}
\begin{aligned}
    \left|B^{T,r,\pi}Q^{\pi}\left(*\right)-B^{\bar{T},\bar{r},\pi}Q^{\pi}\left(*\right)\right| \\
    =\left| (r\left(*\right)-\bar{r}\left(*\right))+\gamma \sum_{s'}\left(T\left(s'|*\right)-\bar{T}\left(s'|*\right)\right)V^{\pi}(s')\right|\\
    \leq \left| (r\left(*\right)-\bar{r}\left(*\right))\right| + \gamma\left| \sum_{s'}\left(T\left(s'|*\right)-\bar{T}\left(s'|*\right)\right)V^{\pi}(s')\right| \\
    \leq \frac{C_{r,\delta}+\gamma C_{T,\delta}2R_{\max}/(1-\gamma)}{\sqrt{|D\left(*\right)|}},
\end{aligned}
\end{equation}
        where $*$ denotes $(s,a)$ and $V^\pi(s) = \mathbb{E}_{a \sim \pi(\cdot|s)} [Q(a|s)] $.
This gives us an expression, which is a function of $C_{r}$ and $C_{T}$, to bound the potential overestimation caused by the sample bias.


In the proposed algorithm CROP, only state transitions of offline data are kept, and the reward for offline data is replaced by the reward of the environment model. Thus, the Bellman operator using offline data is $B^{\bar{T},\hat{r},\pi}$, whose difference to $B^{\bar{T},\bar{r},\pi}$ is bounded by $\epsilon_{r}$:
\begin{equation}
    \begin{split}
        \left|B^{\bar{T},\hat{r},\pi}Q^{\pi}\left(*\right)-B^{\bar{T},\bar{r},\pi}Q^{\pi}\left(*\right)\right| = |\hat{r}\left(*\right)-\bar{r}\left(*\right)| \leq \epsilon_{r}.
    \end{split}
\end{equation}

The Bellman operator using interactions with the model is $B^{\hat{T}, \hat{r}, \pi}$, whose difference to $B^{\bar{T},\bar{r},\pi}$ is bounded as follows:

\begin{equation}
\begin{aligned}
    \left|B^{\hat{T},\hat{r},\pi}Q\left(*\right)-B^{\bar{T},\bar{r},\pi}Q\left(*\right)\right|\\
    = \left|\hat{r}\left(*\right)-\bar{r}\left(*\right)+\gamma \sum_{s'}\left(\hat{T}\left(s'|*\right)-\bar{T}\left(s'|*\right)\right)V^{\pi}(s')\right| \\
    \leq \left|\hat{r}\left(*\right)-\bar{r}\left(*\right)\right|+\left|\gamma \sum_{s'}\left(\hat{T}\left(s'|*\right)-\bar{T}\left(s'|*\right)\right)V^{\pi}(s')\right|\\
    \leq \epsilon_{r}+\gamma \epsilon_{P}2R_{\max}/(1-\gamma).
\end{aligned}
\end{equation}

Since offline data and model-generated interactions are mixed with the ratio $(1-f):f$ for policy optimization in CROP, the Bellman operator used in CROP $B_{\rm{CROP}}$ can be seen as a mix of $B^{\bar{T},\hat{r},\pi}$ and $B^{\hat{T},\hat{r},\pi}$:
\begin{equation}
    \begin{aligned}
    \label{equation: 20}
        B_{\rm{CROP}}Q\left(*\right) = \left[(1-f)B^{\bar{T},\hat{r},\pi} + f B^{\hat{T},\hat{r},\pi}\right]Q\left(*\right)\\
        \leq B^{T,r,\pi}Q\left(*\right) + \left|B^{T,r,\pi}Q\left(*\right)-B^{\bar{T},\bar{r},\pi}Q\left(*\right)\right|\\
        +(1-f)  \left|B^{\bar{T},\hat{r},\pi}Q\left(*\right)-B^{\bar{T},\bar{r},\pi}Q\left(*\right)\right|\\
        +f  \left|B^{\hat{T},\hat{r},\pi}Q\left(*\right)-B^{\bar{T},\bar{r},\pi}Q\left(*\right)\right|\\
        \leq B^{T,r,\pi}Q\left(*\right) + \frac{C_{r,\delta}+\gamma C_{T,\delta}2R_{\max}/(1-\gamma)}{\sqrt{|D\left(*\right)|}}\\
        +(1-f)\epsilon_{r} + f\left(\epsilon_{r}+\gamma \epsilon_{T}2R_{\max}/\left(1-\gamma\right)\right)\\
        = B^{T,R,\pi}Q\left(*\right) - \beta \frac{\mu}{\bar{\pi}(a|s)} + \\\frac{C_{r,\delta}+\gamma C_{T,\delta}2R_{\max}/(1-\gamma)}{\sqrt{|D\left(*\right)|}}
        +\epsilon_{r} + f\gamma \epsilon_{T}2R_{\max}/\left(1-\gamma\right).
    \end{aligned}
\end{equation}

$\hat{Q}^{\pi}(*)$ is the fixed point of $B_{\rm{CROP}}$, and $Q^{\pi}\left(*\right)$ is the fixed point of $B^{T,R,\pi}$. Define the terms independent of $\beta$ in the RHS of Equation~(\ref{equation: 20}) as $\phi$: $\phi = \frac{C_{r,\delta}+\gamma C_{T,\delta}2R_{\max}/(1-\gamma)}{\sqrt{|D\left(s,a\right)|}} + \epsilon_{r} + f*\gamma \epsilon_{T}2R_{\max}/\left(1-\gamma\right)$. 
By computing the fixed point of Equation~(\ref{equation: 20}), $\hat{Q}^{\pi}(*)$ can be bounded as follows:
\begin{equation}
    \label{equation: 21}
    \hat{Q}^{\pi}(*) \leq Q^{\pi}\left(*\right) - \beta*\left[S^{\pi}\frac{\mu}{\bar{\pi}}\right]\left(*\right) + \left[S^{\pi}\phi\right]\left(*\right),
\end{equation}
where $S^{\pi}=\left(I-\gamma T^{\pi}\right)^{-1}$.

Thus, by choosing large enough $\beta$, $\hat{Q}^{\pi}(*) \leq Q^{\pi}\left(*\right)$ and $\mathbb{E}_{s\sim\mu_{0},a\sim\pi(\cdot|s)} \left[\hat{Q}^{\pi}(s,a)\right] \leq \mathbb{E}_{s\sim\mu_{0},a\sim\pi(\cdot|s)} \left[Q^{\pi}(s,a)\right]$.
\end{proof}
Theorem~\ref{theorem: 1} shows that CROP can conservatively estimate Q-function and avoid the common overestimation problem in offline RL. However, not all conservative estimations help avoid OOD actions. As an extreme example, for all bounded $Q^{\pi}$ and $\hat{Q}^{\pi}$, there exists a sufficiently large constant so that $\hat{Q}^{\pi}(*)$ minus the constant is smaller than $Q^{\pi}(*)$. Nevertheless, such constant shift operations on the Q-function preserve the action preference ordering and have no effect on policy optimization. Thus, it is necessary to prove that CROP is effective in avoiding OOD actions.
\begin{theorem}
\label{theorem: 2}
$\forall a_{1}, a_{2} \in A, s_{1}, s_{2} \in S$, if $\bar{\pi}\left(a_{1}|s_{1}\right)<\bar{\pi}\left(a_{2}|s_{2}\right)$, for sufficiently large $\beta$, 
\begin{equation}
    Q^{\pi}\left(s_{1},a_{1}\right) - \hat{Q}^{\pi}(s_{1},a_{1}) > Q^{\pi}\left(s_{2},a_{2}\right) - \hat{Q}^{\pi}(s_{2},a_{2}).
\end{equation}
\end{theorem}
\begin{proof}
Similar to Equation~(\ref{equation: 20}), 
\begin{equation}
    \begin{aligned}
    \label{equation: BQ lowbound}
        B_{\rm{CROP}}Q^{\pi}\left(*\right) = \left[(1-f)B^{\bar{T},\hat{r},\pi} + f B^{\hat{T},\hat{r},\pi}\right]Q^{\pi}\left(*\right)\\
        \geq B^{T,r,\pi}Q^{\pi}\left(*\right) - \left|B^{T,r,\pi}Q^{\pi}\left(*\right)-B^{\bar{T},\bar{r},\pi}Q^{\pi}\left(*\right)\right| \\
        -(1-f) \left|B^{\bar{T},\hat{r},\pi}Q^{\pi}\left(*\right)-B^{\bar{T},\bar{r},\pi}Q^{\pi}\left(*\right)\right|\\
        -f \left|B^{\hat{T},\hat{r},\pi}Q^{\pi}\left(*\right)-B^{\bar{T},\bar{r},\pi}Q^{\pi}\left(*\right)\right|\\
        \geq B^{T,R,\pi}Q^{\pi}\left(*\right) - \beta \frac{\mu}{\bar{\pi}(a|s)} -\phi.
    \end{aligned}
\end{equation}

Computing the fixed points on both sides of Equation~(\ref{equation: BQ lowbound}) yields the following relation between $\hat{Q}^{\pi}$ and $Q^{\pi}$:
\begin{equation}
    \label{equation: 24}
    \hat{Q}^{\pi}(*) \geq Q^{\pi}\left(*\right) - \beta*\left[S^{\pi}\frac{\mu}{\bar{\pi}}\right]\left(*\right) - \left[S^{\pi}\phi\right]\left(*\right)
\end{equation}
Based on Equation~(\ref{equation: 21}) and (\ref{equation: 24}), 
\begin{equation}
\begin{aligned}
    \left(Q^{\pi}\left(s_{1},a_{1}\right) - \hat{Q}^{\pi}(s_{1},a_{1})\right) - \left(Q^{\pi}\left(s_{2},a_{2}\right) - \hat{Q}^{\pi}(s_{2},a_{2})\right) \geq \\
    \beta*\left\{\left[S^{\pi}\frac{\mu}{\bar{\pi}}\right]\left(s_{1},a_{1}\right)-\left[S^{\pi}\frac{\mu}{\bar{\pi}}\right]\left(s_{2},a_{2}\right)\right\}\\ -\left[S^{\pi}\phi\right]\left(s_{1},a_{1}\right)- \left[S^{\pi}\phi\right]\left(s_{2},a_{2}\right)
\end{aligned}
\end{equation}
For large enough $\beta$, $\left(Q^{\pi}\left(s_{1},a_{1}\right) - \hat{Q}^{\pi}(s_{1},a_{1})\right) - \left(Q^{\pi}\left(s_{2},a_{2}\right) - \hat{Q}^{\pi}(s_{2},a_{2})\right) > 0$, i.e., $Q^{\pi}\left(s_{1},a_{1}\right) - \hat{Q}^{\pi}(s_{1},a_{1}) > Q^{\pi}\left(s_{2},a_{2}\right) - \hat{Q}^{\pi}\left(s_{2},a_{2}\right)$ 
\end{proof}
Theorem~\ref{theorem: 2} states that with sufficiently large $\beta$, the conservatism ($Q^{\pi}$ minus $\hat{Q}^{\pi}$) of CROP is stronger for actions that occur less frequently in $\bar{\pi}$, thereby avoiding OOD actions in policy optimization. When $\beta$ comes to $+\infty$, the optimal policy in CROP $\pi^*$ directly selects the most likely action in $\bar{\pi}$ ($\arg\max_a \hat{Q}^{\pi}(\cdot|s) = \arg\max_a \bar{\pi}(\cdot, s)$). An appropriate $\beta$ should be selected in practical applications to balance the conservatism.

\begin{table}[t!]
\begin{center}
\caption{Value of Hyperparameters}
\label{table: hyperparameter}
\begin{threeparttable}
\begin{tabular}{ll}
\hline\hline
\specialrule{0em}{2.5pt}{2.5pt}
\textbf{Hyperparameter} & \textbf{Value}\\
\specialrule{0em}{1.5pt}{1.5pt}
\hline
\specialrule{0em}{2.5pt}{2.5pt}
Hidden units of model & 200 \\ \specialrule{0em}{0.7pt}{0.7pt}
Hidden units of policy & 256 \\ \specialrule{0em}{0.7pt}{0.7pt}
Number of layers in model & 4 \\ \specialrule{0em}{0.7pt}{0.7pt}
Number of layers in Q-function & 2 \\ \specialrule{0em}{0.7pt}{0.7pt}
Number of layers in policy & 2 \\ \specialrule{0em}{0.7pt}{0.7pt}
Ratio of the validation set in $\mathcal{D}$ & 0.01 \\ \specialrule{0em}{0.7pt}{0.7pt}
Nonlinear activation & ReLU \\ \specialrule{0em}{0.7pt}{0.7pt}
Batch size in model training & 256 \\ \specialrule{0em}{0.7pt}{0.7pt}
Batch size in policy optimization & 512 \\ \specialrule{0em}{0.7pt}{0.7pt}
Ratio of model-generated online interactions $f$ & 0.5 \\ \specialrule{0em}{0.7pt}{0.7pt}
The number of random actions $n$ & 10 \\ \specialrule{0em}{0.7pt}{0.7pt}
Optimizer & Adam \\ \specialrule{0em}{0.7pt}{0.7pt}
Learning rate of model & 1e-3\\\specialrule{0em}{0.7pt}{0.7pt}
Learning rate of Q-function & 3e-4\\ \specialrule{0em}{0.7pt}{0.7pt}
Learning rate of policy and $\alpha$ & 1e-4\\ \specialrule{0em}{0.7pt}{0.7pt}
Discount factor $\gamma$ & 0.99 \\ \specialrule{0em}{0.7pt}{0.7pt}
Exponentially moving average of $\bar{Q}$ & 0.005 \\ \specialrule{0em}{0.7pt}{0.7pt}
\hline\hline
\end{tabular}
\end{threeparttable}
\end{center}
\end{table}

Using the conservative reward to mitigate overestimation, the proposed algorithm CROP has a performance lower bound as stated in the following.
\begin{theorem}
\label{theorem: 3}
The optimal policy $\pi^*$ learned by maximizing $\hat{Q}^{\pi}$
\begin{equation}
    \label{equation: optimal policy}
    \pi^* = \arg\max_{\pi} \mathbb{E}_{s,a} \left[\hat{Q}^{\pi}\left(s,a\right)\right]
\end{equation}
has a performance lower bound:
\begin{equation}
    \label{equation: safe policy improvement}
    \begin{aligned}
        \mathbb{E}_{s,a} \left[Q^{\pi^*}\left(s,a\right)\right] \geq 
        \mathbb{E}_{s,a} \left[Q^{\bar{\pi}}\left(s,a\right)\right] + \Delta_{\rm adv} - \delta.
    \end{aligned}
\end{equation}
$\Delta_{\rm adv} = \left(\mathbb{E}_{s,a} \left[\hat{Q}^{\pi^*}(s,a)\right]- \mathbb{E}_{s,a} \left[\hat{Q}^{\bar{\pi}}(s,a)\right]\right)\geq 0$ is the advantage of $\pi^*$ over $\bar{\pi}$ on Equation~(\ref{equation: optimal policy}). $\delta$ is a function of the sample bias, estimation errors, and the difference between $\pi^*$ and $\bar{\pi}$, which is detailed in the proof.
\end{theorem}


\begin{proof}
For ease of writing, we define the expected cumulative reward of policy $\pi$ in an MDP with transition probability $T$ and reward $r$ as $F(T, r, \pi) = \mathbb{E}_{s,a} \left[Q^{\pi}(s,a)\right]$. 
Then $\mathbb{E}_{s,a} \left[Q^{\pi}(s,a)\right] = F(T, r, \pi)$, and
$F(T_f,\hat{r}, \pi)=\mathbb{E}_{s,a} \left[\hat{Q}^{\pi}(s,a)\right]$, where $T_f = (1-f) \bar{T} + f\hat{T}$. $\Delta_{\rm adv} = F\left(T_f,\hat{r}, \pi^*\right) - F\left(T_f,\hat{r}, \bar{\pi}\right)$.

\textbf{Step 1: Relate $F\left(T_f, R, \pi^*\right)$ and $F\left(T_f, R, \bar{\pi}\right)$.} 

Since $\left|\bar{r} - \hat{r}\right|\leq \epsilon_r$,
\begin{equation}
\label{equation: 28}
\begin{aligned}
    F\left(T_f,R - \beta \frac{\mu}{\bar{\pi}\left(a|s\right)} + \frac{C_{r,\delta}}{\sqrt{|D\left(s,a\right)|}} + \epsilon_r, \pi^*\right) \geq F\left(T_f,\hat{r}, \pi^*\right),\\
    F\left(T_f, R - \beta \frac{\mu}{\bar{\pi}(a|s)} - \frac{C_{r,\delta}}{\sqrt{|D\left(s,a\right)|}} - \epsilon_r, \bar{\pi}\right) \leq F\left(T_f,\hat{r}, \bar{\pi}\right).
\end{aligned}
\end{equation}
Rearranging Equation~(\ref{equation: 28}) yields the following expression:
\begin{equation}
\label{equation: 25}
\begin{aligned}
    F\left(T_f,R, \pi^*\right) \geq 
    F\left(T_f, R, \bar{\pi}\right) + F\left(T_f, \beta \frac{\mu}{\bar{\pi}(a|s)}, \pi^*\right) + \Delta_{\rm adv}
    \\-\beta -\frac{2\epsilon_r}{1-\gamma}- F\left(T_f,\frac{C_{r,\delta}}{\sqrt{|D\left(s,a\right)|}},\pi^*\right) 
    \\- F\left(T_f,\frac{C_{r,\delta}}{\sqrt{|D\left(s,a\right)|}},\bar{\pi}\right),
\end{aligned}
\end{equation}
\begin{equation}
\begin{aligned}
    F\left(T_f,\frac{C_{r,\delta}}{\sqrt{|D\left(s,a\right)|}},\pi\right) = \mathbb{E}_{s,a} \left[ \frac{C_{r,\delta}\pi\left(a|s\right) d^{\pi}_{T_f}\left(s\right)}{\sqrt{|D\left(s\right)|\bar{\pi}\left(a|s\right)}} \right]\\
    =\mathbb{E}_{s} \left[\frac{C_{r,\delta}d^{\pi}_{T_f}\left(s\right)}{\sqrt{|D\left(s\right)|}}  \mathbb{E}_{a} \left[ \frac{\pi\left(a|s\right)}{\sqrt{\bar{\pi}\left(a|s\right)}} \right] \right]  \\
    \leq \mathbb{E}_{s} \left[ \frac{C_{r,\delta}d^{\pi}_{T_f}\left(s\right)}{\sqrt{\left|D\left(s\right)\right|}}\sqrt{\left|A\right|\left[D_{\rm{CQL}}\left(\pi,\bar{\pi}\right)\left(s\right)+1\right]} \right],
\end{aligned}
\end{equation}
where $d^{\pi}_{T_f}$ is the distribution of state with the transition probability $T_f$ and policy $\pi$. $D_{\rm{CQL}}(\pi,\bar{\pi})(s)=\mathbb{E}_{a \sim \pi(\cdot|s)} \left[\frac{\pi(a|s)}{\bar{\pi}(a|s)}-1\right]$ is defined as in~\cite{CQL} and can be bounded as:
\begin{equation}
\begin{aligned}
    D_{\rm{CQL}}\left(\pi,\bar{\pi}\right)\left(s\right)+1 \leq \left(\mathbb{E}_a \left[ \frac{\pi\left(a|s\right)}{\sqrt{\bar{\pi}\left(a|s\right)}}\right]\right)^2,\\
    \left|A\right|\left[ D_{\rm{CQL}}\left(\pi,\bar{\pi}\right)\left(s\right)+1\right] \geq \left(\mathbb{E}_a \left[ \frac{\pi\left(a|s\right)}{\sqrt{\bar{\pi}\left(a|s\right)}}\right]\right)^2.
\end{aligned}
\end{equation}
\begin{equation}
\begin{aligned}
    F\left(T_f,\beta \frac{\mu}{\bar{\pi}\left(a|s\right)},\pi\right) = \mathbb{E}_{s,a}\left[\beta \frac{\mu}{\bar{\pi}\left(a|s\right)} \pi\left(a|s\right) d^{\pi}_{T_f}\left(s\right)\right]\\
    = \mathbb{E}_{s}\left[\beta d^{\pi}_{T_f}\left(s\right) \mathbb{E}_{a}\left[\frac{\pi\left(a|s\right)}{\sqrt{\bar{\pi}\left(a|s\right)}}\frac{\mu}{\sqrt{\bar{\pi}\left(a|s\right)}}\right]  \right] \\
    \geq \mathbb{E}_{s}\left[\beta d^{\pi}_{T_f}\left(s\right)\sqrt{D_{\rm{CQL}}\left(\pi,\bar{\pi}\right)\left(s\right)+1}\sqrt{D_{\rm{CQL}}\left(\pi_{\rm{ran}},\bar{\pi}\right)\left(s\right)+1}\right],
\end{aligned}
\end{equation}
where $\pi_{\rm{ran}}$ is the random policy and $\pi_{\rm{ran}}(s,a) = \mu$ for all $(s,a)$.

Therefore, 
\begin{equation}
    \begin{aligned}
        F\left(T_f,R, \pi^*\right) \geq F\left(T_f, R, \bar{\pi}\right) + \Delta_{\rm adv} - \Delta_R,
    \end{aligned}
\end{equation}
where 
\begin{equation}
    \begin{aligned}
    \Delta_R = 
    \mathbb{E}_{s}\left[\frac{C_{r,\delta}}{\sqrt{\left|D\left(s\right)\right|}}\left\{d^{\pi^*}_{T_f}\left(s\right)\sqrt{\left|A\right|\left[D_{\rm{CQL}}\left(\pi^*,\bar{\pi}\right)\left(s\right) +1\right]} \right. \right.
    \\ \left.\left.+d^{\bar{\pi}}_{T_f}\left(s\right)\sqrt{\left|A\right|}\right\}\right] + \beta +\frac{2\epsilon_r}{1-\gamma}
    \\-\mathbb{E}_s \left[\beta d^{\pi}_{T_f}\left(s\right)\sqrt{D_{\rm{CQL}}\left(\pi,\bar{\pi}\right)\left(s\right)+1}
    \sqrt{D_{\rm{CQL}}(\pi_{\rm{ran}},\bar{\pi})\left(s\right)+1}\right]
    \end{aligned}
\end{equation}

\begin{figure}[t]
\centering
\includegraphics[width=\columnwidth]{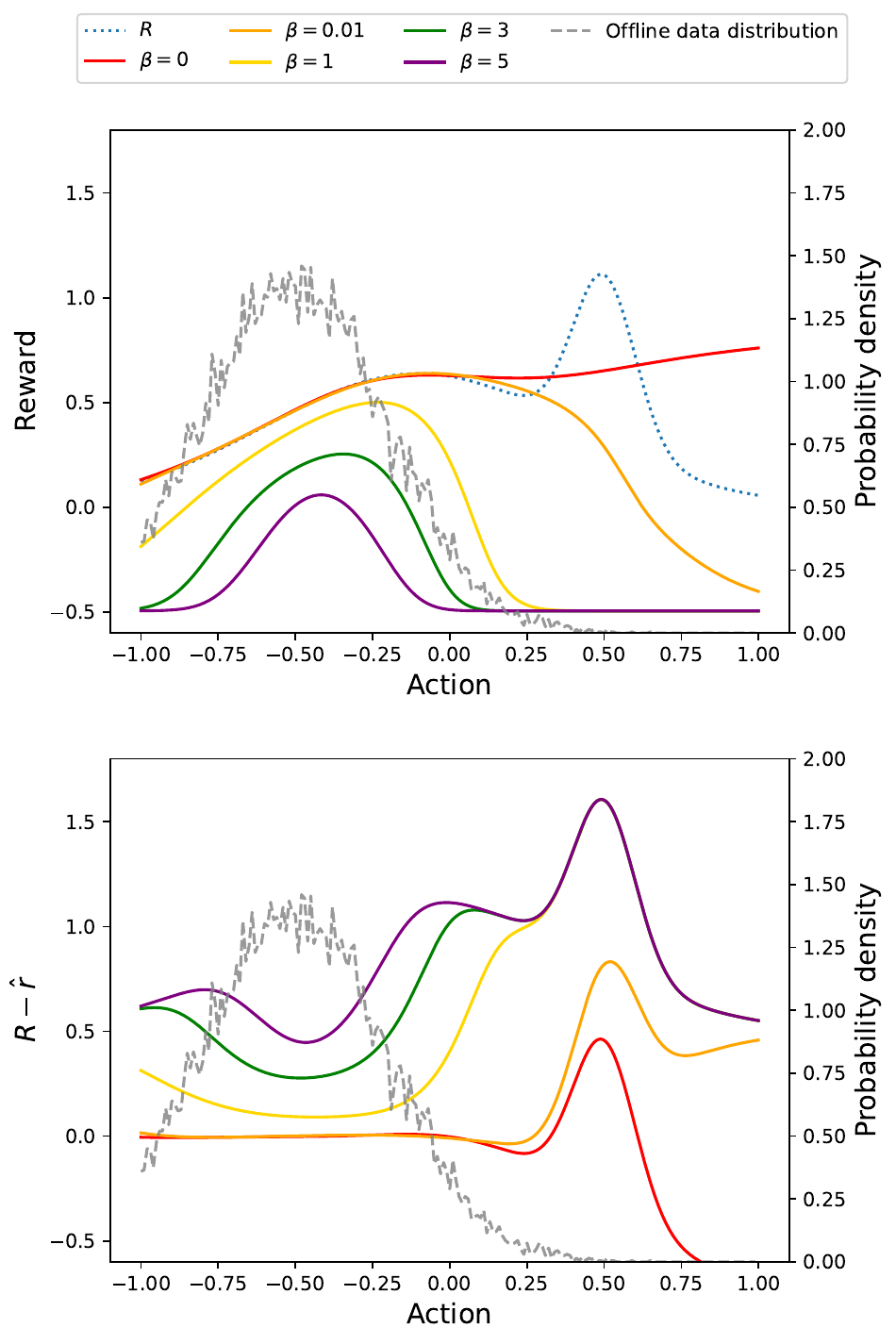} 
\caption{Conservative reward $\hat{r}$ when offline data is sampled with normal distribution. $R$ is also shown for comparison.}
\label{fig: normal distribution}
\end{figure}
\begin{figure}[t]
\centering
\includegraphics[width=\columnwidth]{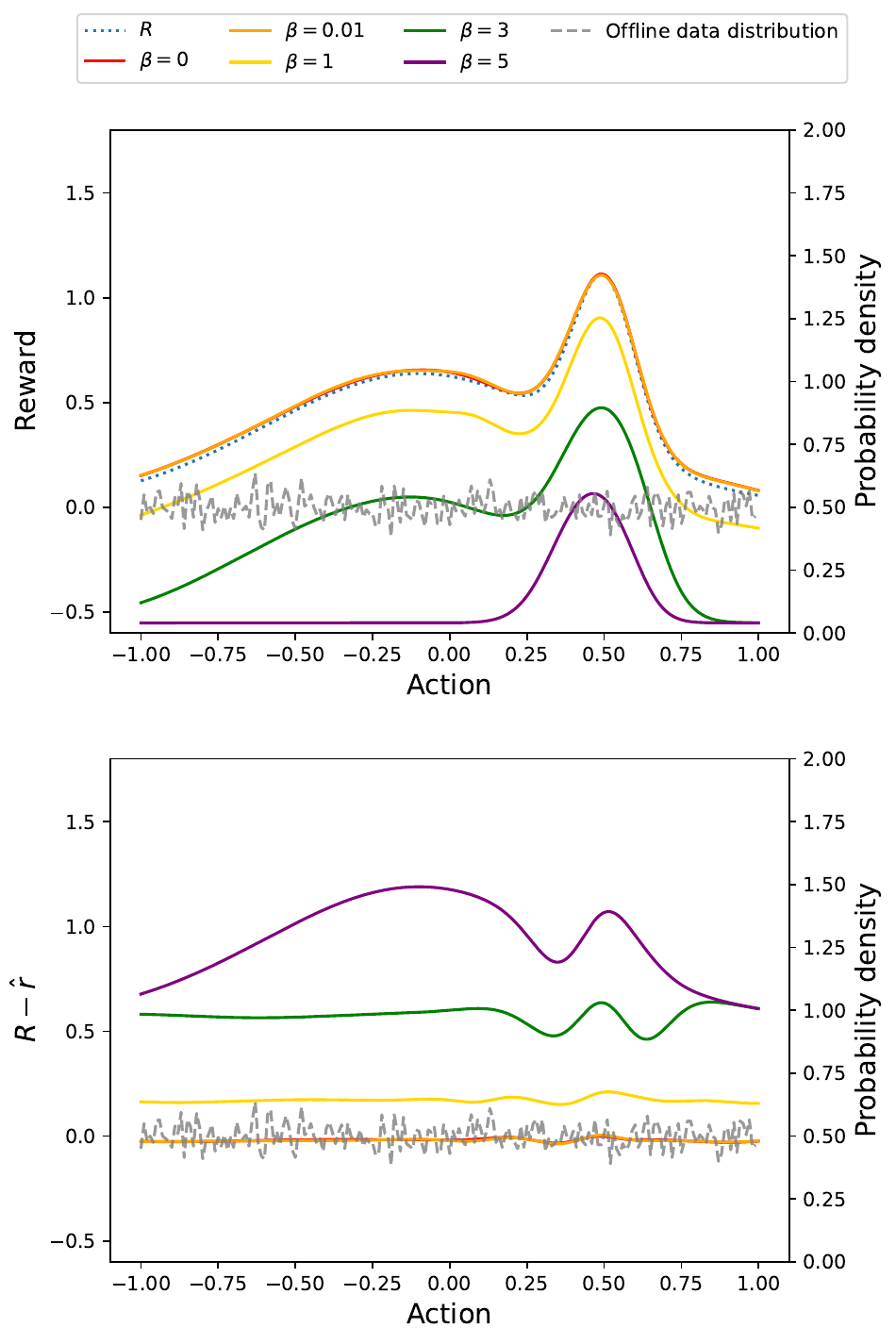} 
\caption{Conservative reward $\hat{r}$ when offline data is sampled with uniform distribution.}
\label{fig: uniform distribution}
\end{figure}

\textbf{Step 2: Relate $F\left(T_f, R, \cdot\right)$ and $F\left(T, R, \cdot\right)$.}


\begin{equation}
    \begin{aligned}
        F\left(T_f, R, \pi\right) = F\left(T, R, \pi\right) + \\\frac{\gamma }{1-\gamma}\mathbb{E}_{s,a\sim d_{T_f}^{\pi}\left(s\right)\pi\left(a|s\right)}\left[(T_f^{\pi}-T^{\pi})Q^{\pi}\right]
    \end{aligned}
\end{equation}
The above equation comes from the Simulation Lemma (Chapter 2
, Lemma 2.2) in \cite{RLTheory}. Therefore,
\begin{equation}
    \begin{aligned}
        &\left|F\left(T_f, R, \pi\right) - F\left(T, R, \pi\right)\right| \\
        &\leq\frac{\gamma R_{\max} }{(1-\gamma)^2}\mathbb{E}_{s,a\sim d_{T_f}^{\pi}\left(s\right)\pi\left(a|s\right)}\left[||T_f^{\pi}-T^{\pi}||_{1}\right]\\
        &\leq \frac{\gamma R_{\max}}{\left(1-\gamma\right)^2}\left\{\mathbb{E}_{s,a\sim d_{T_f}^{\pi}\left(s\right)\pi\left(a|s\right)}\left[||\bar{T}^{\pi}-T^{\pi}||_1\right] + f\epsilon_P\right\}
    \end{aligned}
\end{equation}
\begin{equation}
    \begin{aligned}
        &\mathbb{E}_{s,a\sim d_{T_f}^{\pi}\left(s\right)\pi\left(a|s\right)}\left[||\bar{T}^{\pi}-T^{\pi}||_1\right] \\
        &= \mathbb{E}_{s,a} \left[ ||\bar{T}\left(\cdot|s,a\right)-T\left(\cdot|s,a\right)||_1\pi\left(a|s\right)d^{\pi}_{T_f}\left(s\right)\right] \\
        &\leq \mathbb{E}_{s,a} \left[ \frac{C_{T,\delta}}{\sqrt{\left|D\left(s,a\right)\right|}}\pi\left(a|s\right)d^{\pi}_{T_f}\left(s\right)\right]\\
        &= \mathbb{E}_{s} \left[\frac{C_{T,\delta}d^{\pi}_{T_f} \left(s\right)}{\sqrt{\left|D\left(s\right)\right|}}\mathbb{E}_{a} \left[\frac{\pi\left(a|s\right)}{\sqrt{\bar{\pi}\left(a|s\right)}} \right] \right] \\
        &\leq \mathbb{E}_{s} \left[\frac{C_{T,\delta}d^{\pi}_{T_f} \left(s\right)}{\sqrt{\left|D\left(s\right)\right|}}\sqrt{|A|\left[D_{\rm{CQL}}\left(\pi,\bar{\pi}\right)\left(s\right)+1\right]} \right]
    \end{aligned}
\end{equation}
so that
\begin{equation}
    \begin{aligned}
        &\left|F\left(P_f, R, \pi\right) - F\left(P, R, \pi\right)\right| \leq \Delta_P\left(\pi\right) =\\
        &\frac{\gamma R_{\max}}{\left(1-\gamma\right)^2}\left\{f\epsilon_P+\mathbb{E}_s \left[\frac{C_{T,\delta}d^{\pi}_{T_f} (s)}{\sqrt{\left|D\left(s\right)\right|}}\sqrt{\left|A\right|\left[D_{\rm{CQL}}\left(\pi,\bar{\pi}\right)\left(s\right)+1\right]}\right]\right\} \\
    \end{aligned}
\end{equation}

\textbf{Step 3: Relate $F\left(P, R, \pi^*\right)$ and $F\left(P, R, \bar{\pi}\right)$.}

Combining step 1 and step 2,
\begin{equation}
    \begin{aligned}
        F\left(P, R, \pi^*\right) &\geq F\left(P, R, \bar{\pi}\right) + \Delta_{\rm adv} - \delta\\
        & =F\left(P, R, \bar{\pi}\right) + F\left(T_f,\hat{r}, \pi^*\right) - F\left(T_f,\hat{r}, \bar{\pi}\right) - \delta
    \end{aligned}
\end{equation}
where $\delta = \Delta_R + \Delta_P\left(\pi^*\right) + \Delta_P\left(\bar{\pi}\right)$.
\end{proof}

\section{Experiment}
\label{section: experiment}
\subsection{Conservative reward visualization}
\label{subsection: visualization}
To visualize the proposed conservative reward, we design a simple 1D MDP, where the state is always 0 and the action space is $[-1,1]$. The reward $R$ is defined as:
\begin{equation}
    R(a) = 0.2 * \mathcal{N}\left(a|0.5,0.1\right) + 0.8 * \mathcal{N}\left(a|-0.1,0.5\right) + 0.2 * \epsilon,
\end{equation}
where $\mathcal{N}(*|x,y)$ denotes the probability density function of a Gaussian distribution with mean $x$ and variance $y$.
The noise term $\epsilon$ follows the standard Gaussian distribution.
The offline dataset comprises 20,000 interactions, where actions follow a Gaussian distribution $\mathcal{N}(\cdot|-0.5, 0.3)$ and a uniform distribution $U(-1,1)$, respectively. The model's hyperparameters align with those presented in Table~\ref{table: hyperparameter}.

Fig.~\ref{fig: normal distribution} illustrates the conservative reward for varying values of $\beta$ with a normal data distribution.
Employing the traditional approach of minimizing estimation error ($\beta=0$) for reward estimation, we observe that while the estimates are accurate in data-rich regions ($a<0$), a substantial estimation error exists in data-scarce regions ($a>0.25$). This error may accumulate through bootstrapping, potentially leading to catastrophic overestimation, as discussed in~\cite{CQL}.
As $\beta$ increases, the estimated reward diminishes, indicating heightened conservatism. Furthermore, the smaller $\bar{\pi}(*)$ indicates a more conservative $\hat{r}(*)$. This trend is more pronounced at higher $\beta$ values.
Fig.~\ref{fig: uniform distribution} shows the situation that offline data are sampled from a uniform distribution. When $\beta$ is 0.01, 1, and 3, the differences in conservatism ($R-\hat{r}$) between actions are significantly smaller than in the case with a normal distribution. When $\beta=5$, the conservatism differences between actions are more pronounced than when $\beta$ is smaller. This is because $\hat{r}$ for some actions have reached the estimated lower bound.
The above visualization results prove our theory that the conservative reward estimation we proposed can more strongly underestimate out-of-distribution actions than in-distribution actions, thereby alleviating the distribution shift problem.





\subsection{Experiments on D4RL}
\label{subsection: Experiments on D4RL}

\begin{table*}[]
\label{table:}
\begin{center}
\caption{Value of $\beta$ and $k$ for each dataset. ``R'', ``M-R'', ``M'', and ``M-E'' denote ``Random'', ``Medium-Replay'', ``Medium'', and ``Medium-Expert'' dataset, respectively.}
\label{table: beta and k}
\begin{tabular}{lllllllllllll}
\hline\hline
\specialrule{0em}{2.5pt}{2.5pt}
\multirow{2}{*}{Hyperparameter} & \multicolumn{4}{c}{Halfcheetah}              & \multicolumn{4}{c}{Hopper}                   & \multicolumn{4}{c}{Walker2d}                 \\ \specialrule{0em}{0.7pt}{0.7pt}
\cmidrule(ll){2-5} \cmidrule(ll){6-9} \cmidrule(ll){10-13}
                                         & \multicolumn{1}{c}{R} & \multicolumn{1}{c}{M-R} & \multicolumn{1}{c}{M} & \multicolumn{1}{c}{M-E} & \multicolumn{1}{c}{R} & \multicolumn{1}{c}{M-R} & \multicolumn{1}{c}{M} & \multicolumn{1}{c}{M-E} & \multicolumn{1}{c}{R} & \multicolumn{1}{c}{M-R} & \multicolumn{1}{c}{M} & \multicolumn{1}{c}{M-E} \\
\specialrule{0em}{1.5pt}{1.5pt}
\hline
\specialrule{0em}{2.5pt}{2.5pt}
$\beta$           & 0.01       & 0.01         & 0.001      & 0.001        & 0.01       & 0.01         & 0.05       & 0.05         & 0.01       & 0.01         & 0.05       & 0.05         \\ \specialrule{0em}{0.7pt}{0.7pt}
$k$               & 5          & 5            & 5          & 5            & 5          & 5            & 5          & 10           & 10         & 10           & 10         & 10 \\ \specialrule{0em}{0.7pt}{0.7pt}
\hline\hline        
\end{tabular}
\end{center}
\end{table*}

\begin{table*}[]
    \centering
    \caption{Normalized score on the Mujoco tasks of D4RL dataset}
    \label{table: results on D4RL mujoco}
    \begin{threeparttable}
    \begin{tabular}{ccccccccccc}
    \hline\hline
    \specialrule{0em}{2.5pt}{2.5pt}
    \multicolumn{1}{c}{\multirow{3.5}{*}{Dataset}} & \multicolumn{6}{c}{Model-based methods} & \multicolumn{3}{c}{Model-free methods} \\
    \cmidrule(ll){2-7} \cmidrule(ll){8-10}
    \specialrule{0em}{0.7pt}{0.7pt}
    \multicolumn{1}{c}{} & CROP & \multirow{2}{*}{COMBO} & \multirow{2}{*}{RAMBO} & \multirow{2}{*}{PMDB} & CABI+ & Count- & \multirow{2}{*}{ATAC} & \multirow{2}{*}{EDAC} & \multirow{2}{*}{IQL}\\
    \multicolumn{1}{c}{} & (ours) &  &  &  & TD3-BC & MORL &  &  & \\
    \specialrule{0em}{1.5pt}{1.5pt}
    \hline
    \specialrule{0em}{2.5pt}{2.5pt}
    Halfcheetah-R&33.7$\pm$2.3&\textbf{38.8}&\textbf{40.0}&\textbf{37.8}&15.1&\underline{\textbf{41.0}}&3.9&28.4&-\\ \specialrule{0em}{0.7pt}{0.7pt}
    Hopper-R&\textbf{31.8$\pm$0.4}&7.0&11.5&\underline{\textbf{32.7}}&11.9&\textbf{30.7}&17.5&25.3&- \\ \specialrule{0em}{0.7pt}{0.7pt}
    Walker2d-R&\textbf{20.9$\pm$0.8}&17.9&\textbf{21.6}&\textbf{21.8}&6.4&\underline{\textbf{21.9}}&6.8&16.6&- \\ \specialrule{0em}{0.7pt}{0.7pt}
    Halfcheetah-M-R&\textbf{70.4$\pm$1.6}&55.1&\textbf{68.9}&\underline{\textbf{71.7}}&44.4&\textbf{71.5}&48.0&61.3&44.2 \\ \specialrule{0em}{0.7pt}{0.7pt}
    Hopper-M-R&\textbf{99.4$\pm$4.3}&89.5&\textbf{96.6}&\underline{\textbf{106.2}}&31.3&\textbf{101.7}&\underline{\textbf{102.5}}&\textbf{101.0}&\textbf{94.7} \\ \specialrule{0em}{0.7pt}{0.7pt}
    Walker2d-M-R&\textbf{91.0$\pm$0.5}&56.0&\textbf{85.0}&79.9&29.4&\textbf{87.7}&\underline{\textbf{92.5}}&\textbf{87.1}&73.9 \\ \specialrule{0em}{0.7pt}{0.7pt}
    Halfcheetah-M&\textbf{74.1$\pm$0.8}&54.2&\underline{\textbf{77.6}}&\textbf{75.6}&45.1&\textbf{76.5}&53.3&65.9&47.4 \\ \specialrule{0em}{0.7pt}{0.7pt}
    Hopper-M&\textbf{102.1$\pm$1.3}&\textbf{97.2}&92.8&\underline{\textbf{106.8}}&\textbf{105.0}&\textbf{103.6}&85.6&\textbf{101.6}&66.3 \\ \specialrule{0em}{0.7pt}{0.7pt}
    Walker2d-M&\underline{\textbf{95.4$\pm$1.3}}&\textbf{81.9}&\textbf{86.9}&\textbf{94.2}&82.0&\textbf{87.6}&\textbf{89.6}&\textbf{92.5}&78.3 \\ \specialrule{0em}{0.7pt}{0.7pt}
    Halfcheetah-M-E&\textbf{104.9$\pm$1.9}&90.0&93.7&\underline{\textbf{108.5}}&\textbf{107.6}&\textbf{100.0}&94.8&\textbf{106.3}&86.7 \\ \specialrule{0em}{0.7pt}{0.7pt}
    Hopper-M-E&\textbf{107.5$\pm$3.0}&\textbf{111.1}&83.3&\textbf{111.8}&\underline{\textbf{112.4}}&\textbf{111.4}&\textbf{111.9}&\textbf{110.7}&91.5 \\ \specialrule{0em}{0.7pt}{0.7pt}
    Walker2d-M-E&\textbf{111.6$\pm$1.7}&\textbf{103.3}&68.3&\textbf{111.9}&\textbf{108.6}&\textbf{112.3}&\textbf{114.2}&\underline{\textbf{114.7}}&\textbf{109.6} \\ \specialrule{0em}{0.7pt}{0.7pt}
    \hline
    \specialrule{0em}{0.7pt}{0.7pt}
    Mean&\textbf{78.6$\pm$1.7}&68.9&65.7&\underline{\textbf{79.9}}&58.3&\textbf{78.8}&68.4&\textbf{76.0}&-\\ \specialrule{0em}{0.7pt}{0.7pt}
    \hline\hline
    \end{tabular}
    \begin{tablenotes}
    \item[*] The highest score on each dataset is underlined. Boldface denotes performance better than 90\% of the highest score.
    \end{tablenotes}
    \end{threeparttable}
\end{table*}


\begin{figure}[t!]
    \centering
    \subfigure[]{
    \includegraphics[width=0.28\linewidth]{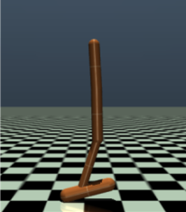}
    \label{subfig: hopper}
    }
    \subfigure[]{
    \includegraphics[width=0.28\linewidth]{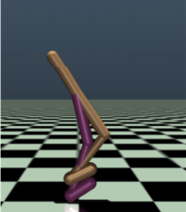}
    \label{subfig: walker2d}
    }
    \subfigure[]{
    \includegraphics[width=0.28\linewidth]{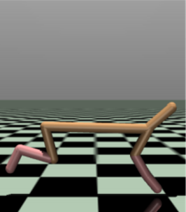}
    \label{subfig: halfcheetah}
    }
    \caption{The Mujoco-v2 tasks. (a) Hopper. (b) Walker2d. (c) Halfcheetah.}
    \label{fig: environment}
\end{figure}

In this section, the proposed method CROP is compared with several representative offline RL methods on the Mujoco-v2 tasks (Hopper, Walker2D, HalfCheetah) of D4RL dataset~\cite{d4rl}.
Each task has four datasets, ``Random'', ``Medium'', ``Medium-Replay'', and ``Medium-Expert''. The ``Random'' dataset comprises transitions gathered through a random policy. The ``Medium'' dataset consists of suboptimal data collected by an early-stopped SAC policy. The ``Medium-Replay'' dataset encompasses the replay buffer generated during the training of an early-stopped SAC policy. Lastly, the ``Medium-Expert'' dataset combines expert demonstrations and suboptimal data.

The heterogeneity in dataset scales and behavior policies leads to varying degrees of data coverage and model accuracy, necessitating dataset-specific tuning of the conservatism coefficient $\beta$ and rollout horizon $k$.
For each dataset, $\beta$ is searched from $\{0.001, 0.01, 0.05\}$ and $k$ is searched from $\{5,10\}$. Details of $\beta$ and $k$ are shown in Table~\ref{table: beta and k}.
We train an ensemble of seven models and select the best five models based on their loss on the validation set. For the Hopper task, the hidden units of Q-function are 256, whereas the hidden units of Q-function are 512 for the Halfcheetah task and the Walker2d task.
Other hyperparameters are shown in Table~\ref{table: hyperparameter}.

The performance is compared with several state-of-the-art model-based (COMBO~\cite{COMBO}, RAMBO~\cite{RAMBO}, PMDB~\cite{PMDB}, CABI+TD3-BC~\cite{CABI}
and Count-MORL~\cite{count}) and model-free (ATAC~\cite{ATAC}, EDAC~\cite{EDAC}, and IQL~\cite{IQL}) offline RL methods. Results of the baselines are obtained from their respective papers.
The score of CROP is the average of the last five evaluations, each of which consists of fifty episodes. The experiments are performed on five random seeds.

\begin{figure}
    \centering
    \includegraphics[width=\linewidth]{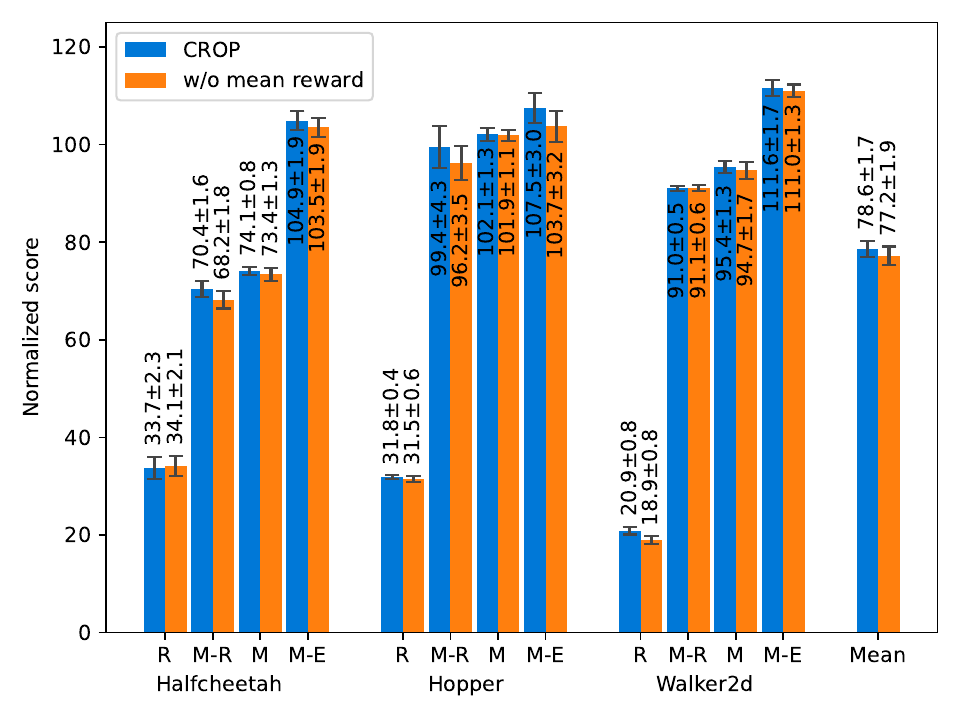}
    \caption{Results of ablation experiments.}
    \label{fig: ablation experiment}
\end{figure}

\begin{figure*}[h]
    \centering
    \includegraphics[width=0.32\linewidth]{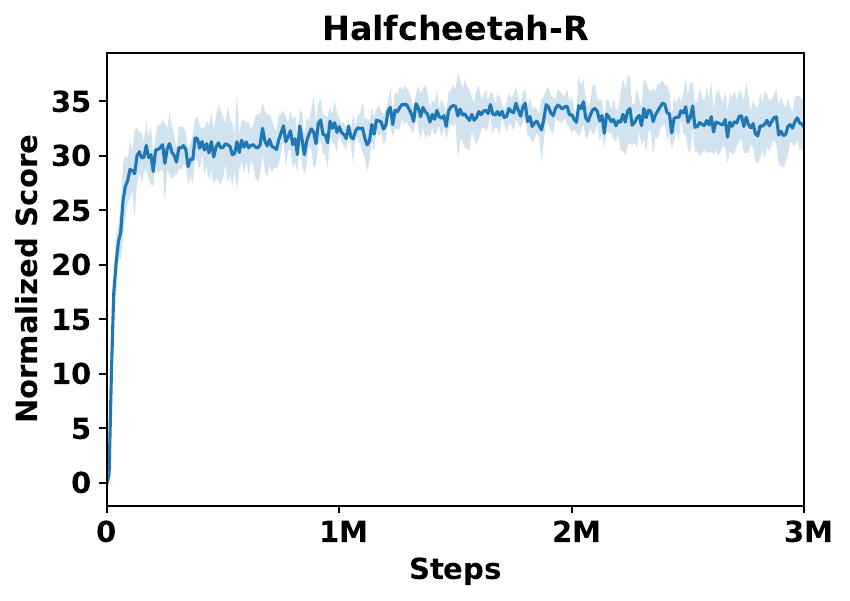}
    \includegraphics[width=0.32\linewidth]{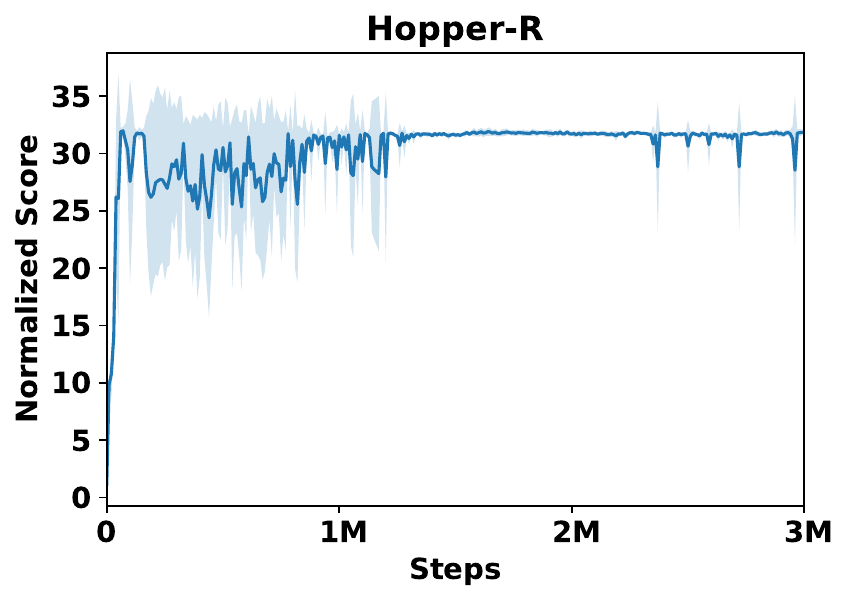}
    \includegraphics[width=0.32\linewidth]{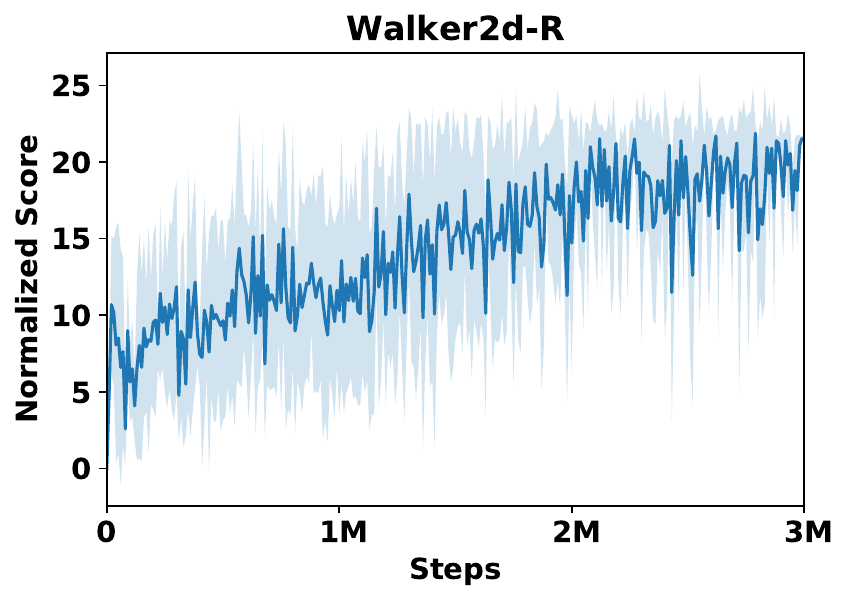}
    \includegraphics[width=0.32\linewidth]{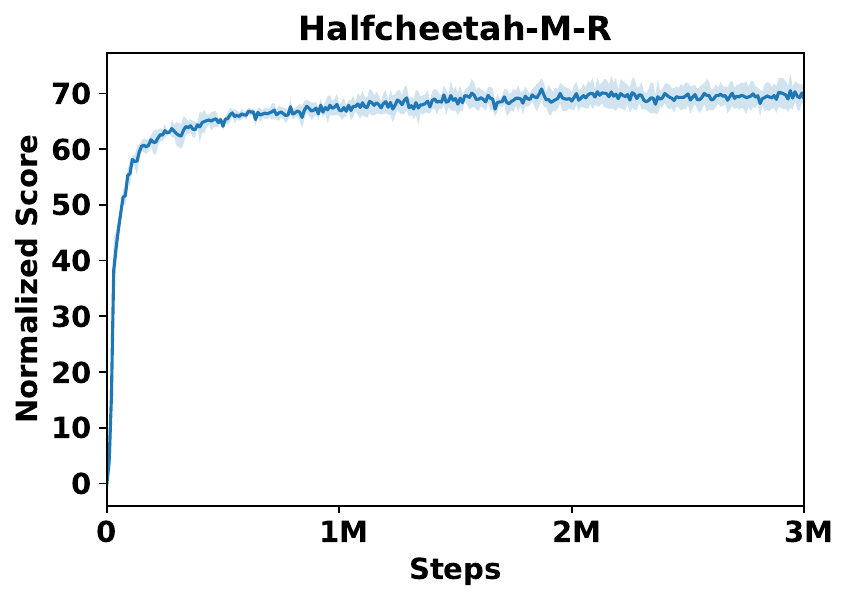}
    \includegraphics[width=0.32\linewidth]{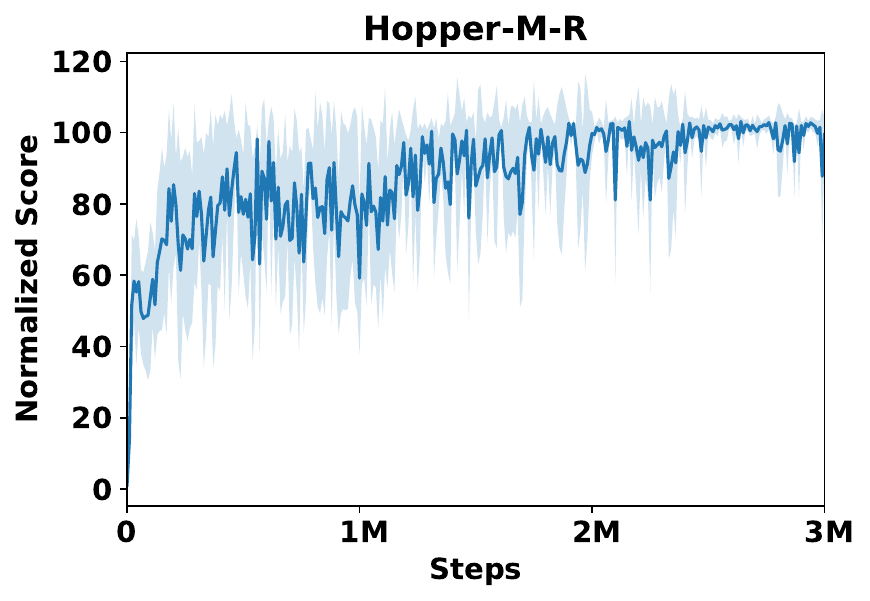}
    \includegraphics[width=0.32\linewidth]{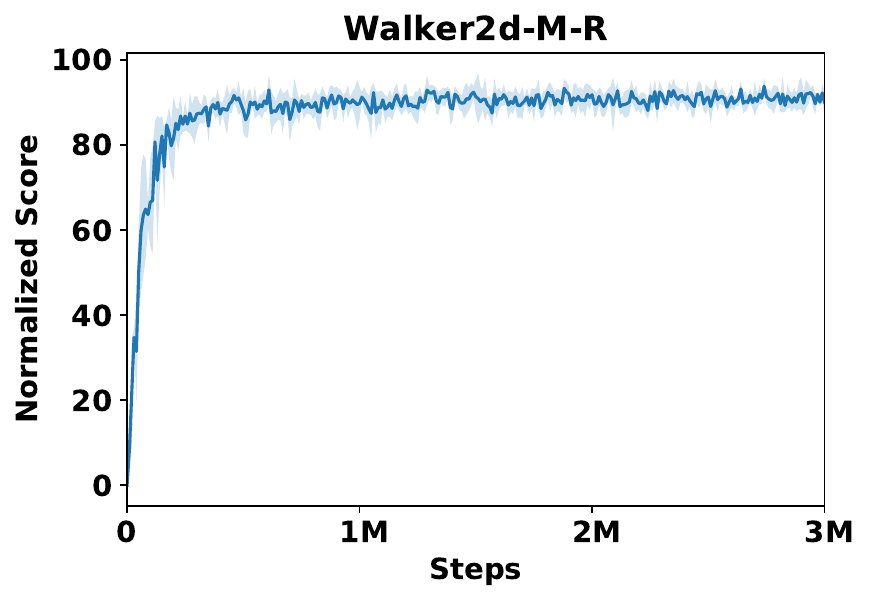}
    \includegraphics[width=0.32\linewidth]{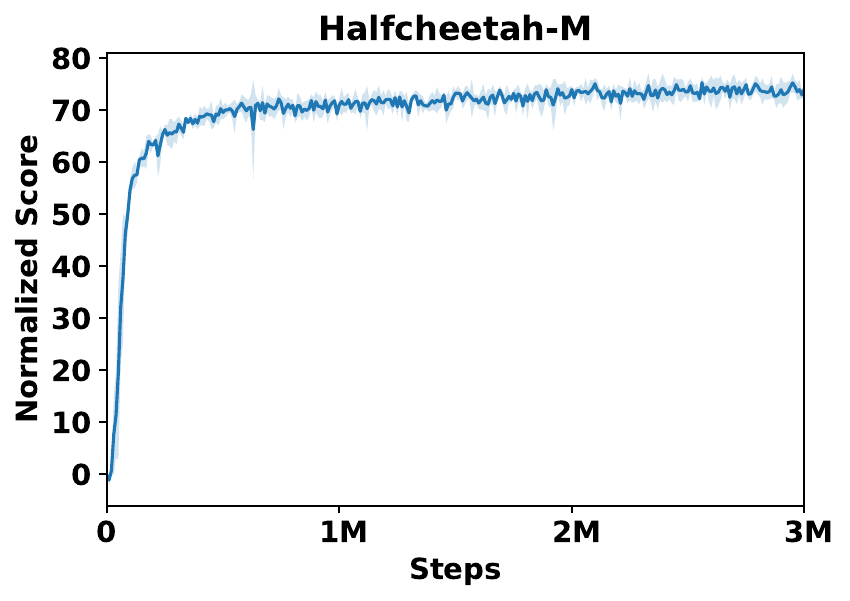}
    \includegraphics[width=0.32\linewidth]{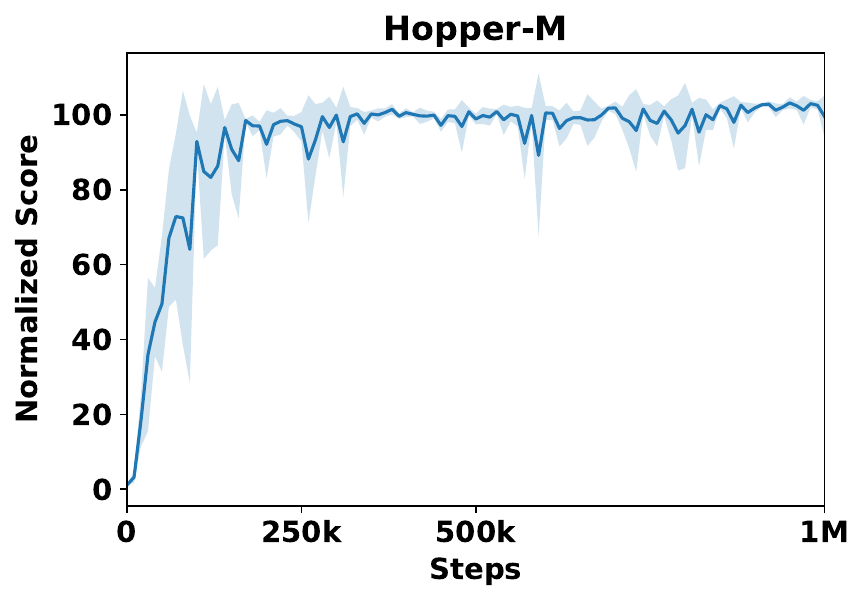}
    \includegraphics[width=0.32\linewidth]{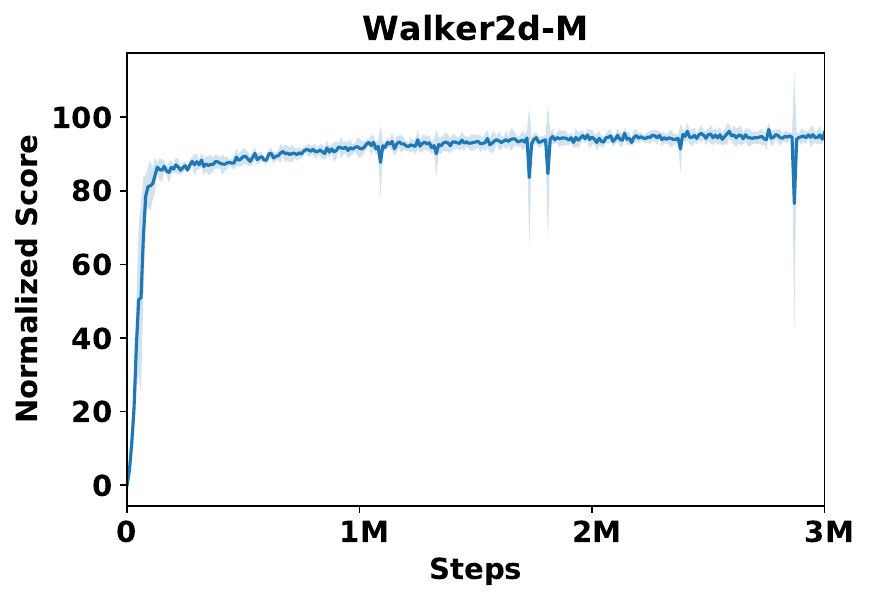}
    \includegraphics[width=0.32\linewidth]{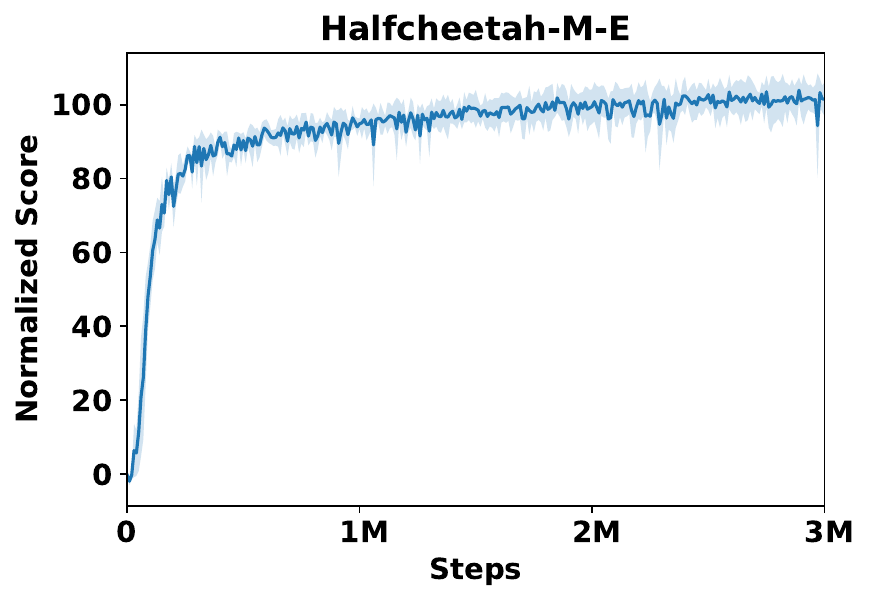}
    \includegraphics[width=0.32\linewidth]{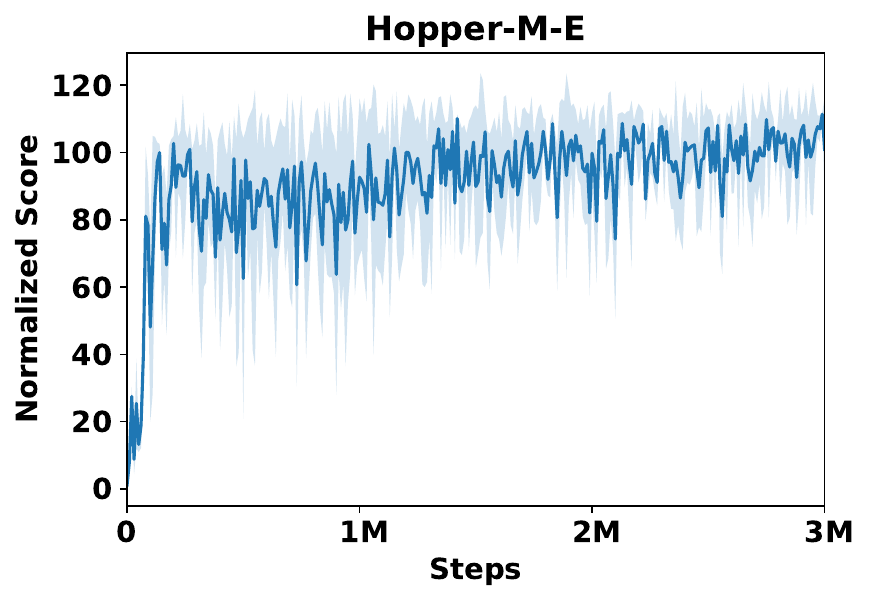}
    \includegraphics[width=0.32\linewidth]{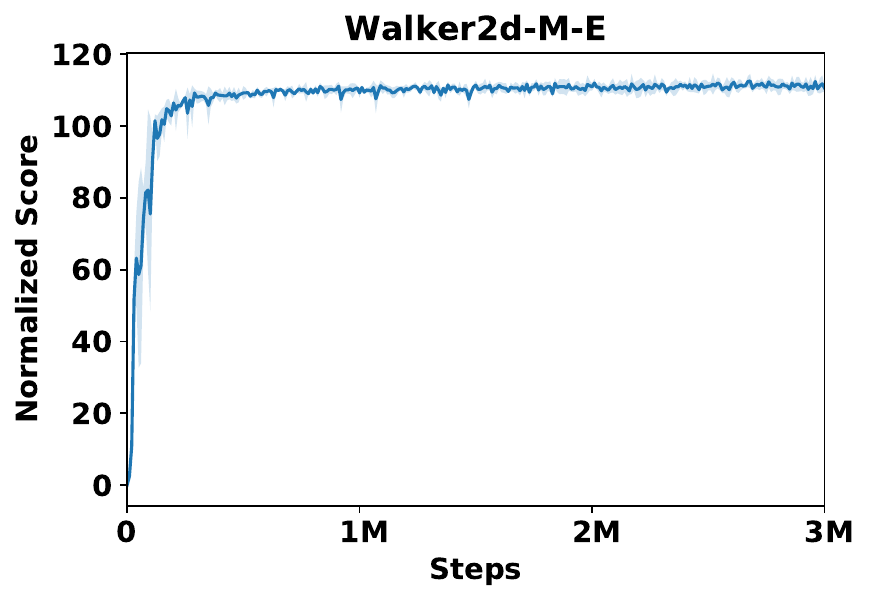}
    \caption{Normalized score during training.}
    \label{fig: learning curve}
\end{figure*}

The mean and standard deviation of normalized scores for CROP and baselines are presented in Table~\ref{table: results on D4RL mujoco}. The performance curve during training is shown in Fig.~\ref{fig: learning curve}.
CROP demonstrates consistent and competitive performance on 11 out of 12 datasets, with a mean score of 78.6. The proposed method performs closely to Count-MORL and PMDB, and performs favorably against several other baseline methods.
Notably, CROP outperforms methods that integrate conservatism into the value function update (e.g., COMBO) or the entire environment model (e.g., RAMBO), highlighting the efficacy of our novel design in introducing conservatism into the reward estimator update. More importantly, CROP achieves a performance comparable to Count-MORL with a markedly simpler design. Count-MORL utilizes an auxiliary state-action frequency estimator to apply a direct penalty inversely proportional to the frequency square root, while CROP simply minimizes the reward for random actions during the reward update to introduce an implicit penalty inversely proportional to the action frequency.

We conducted independent experiments using five different random seeds in all D4RL tasks and statistically analyzed the mean and standard deviation of the results of the last five evaluations in each group of experiments. The standard deviation ranges of most tasks are between $\pm 0.5$ and $\pm 3.0$, showing good convergence and stability. This result indicates that, despite the uncertainties of random initialization, fluctuations during model training, and data sampling, CROP can still maintain consistent performance, reflecting that its model structure design and conservative reward mechanism have strong anti-disturbance capabilities.

\begin{table*}[h!]
\begin{center}
\caption{Normalized score for each dataset with different random action number $n$.}
\label{table: n for each dataset}
\begin{tabular}{cccccccccccccc}
\hline\hline
\specialrule{0em}{2.5pt}{2.5pt}
\multirow{2}{*}{$n$} & \multicolumn{4}{c}{Halfcheetah}              & \multicolumn{4}{c}{Hopper}                   & \multicolumn{4}{c}{Walker2d}                & \multirow{2}{*}{Mean} \\ \specialrule{0em}{0.7pt}{0.7pt}
\cmidrule(ll){2-5} \cmidrule(ll){6-9} \cmidrule(ll){10-13}
  & \multicolumn{1}{c}{R} & \multicolumn{1}{c}{M-R} & \multicolumn{1}{c}{M} & \multicolumn{1}{c}{M-E} & \multicolumn{1}{c}{R} & \multicolumn{1}{c}{M-R} & \multicolumn{1}{c}{M} & \multicolumn{1}{c}{M-E} & \multicolumn{1}{c}{R} & \multicolumn{1}{c}{M-R} & \multicolumn{1}{c}{M} & \multicolumn{1}{c}{M-E}\\
\specialrule{0em}{1.5pt}{1.5pt}
\hline
\specialrule{0em}{2.5pt}{2.5pt}
5 & 32.0 & 67.8 & 74.2 & 102.9 & 31.6 & 93.1 & 100.9 & 100.5 & 21.6 & 90.2 & 94.8 & 110.7 & 76.7 \\
10           & 33.7       & 70.4         & 74.1      & 104.9        & 31.8       & 99.4         & 102.1       & 107.5         & 20.9       & 91.0         & 95.4       & 111.6    & 78.6    \\ \specialrule{0em}{0.7pt}{0.7pt}
50 & 33.7 & 69.3 & 74.8 & 105.4 & 31.6 & 99.1 & 101.2 & 105.1 & 21.8 & 90.3 & 96.4 & 112.3 & 78.4\\
\specialrule{0em}{0.7pt}{0.7pt}
\hline\hline        
\end{tabular}
\end{center}
\end{table*}

\subsection{Ablation experiment}

CROP calculates the estimated reward by taking the mean of $\hat{r}$, in contrast to most existing algorithms that estimate rewards based on a model randomly chosen from an ensemble. We conducted ablation experiments to evaluate the impact of this trick. The hyperparameter configurations are maintained as outlined in Section~\ref{subsection: Experiments on D4RL}. 
The results are presented in Fig.~\ref{fig: ablation experiment}. The results suggest that the performance of CROP mainly comes from the novel conservative reward estimation, while utilizing the ensemble mean for reward estimation, rather than a stochastic selection from the ensemble, yields marginal performance gains (78.6 over 77.2). 
This enhancement aligns with the theoretical framework of Equation~(\ref{equation: q loss}), which operates on reward expectations - the ensemble mean naturally reduces estimation variance compared to individual models.
This trick has the potential to be combined with other model-based offline RL methods to improve performance and stability.

The conservative reward estimation method designed by CROP, i.e., Equation~(\ref{equation: reward loss}), requires a new hyperparameter $n$, which represents the number of random actions used in Equation~(\ref{equation: reward loss}). Experiments were conducted to determine the optimal hyperparameter $n$. The overall mean normalized score is depicted in Table~\ref{table: n for each dataset}. Our analysis indicates that a lower value of $n$ results in a noticeable decrease in performance (76.7 when $n=5$). This reduction in efficacy is likely due to variations in the estimated mean of $\hat{r}(s,\bar{a})$. On the other hand, for larger values of $n$, the difference in performance becomes minimal (78.6 and 78.4 when $n=10$ and $n=50$). 
These results indicate that CROP exhibits robust performance with respect to the hyperparameter n, demonstrating insensitivity beyond a critical threshold value.

\section{Discussion}
\label{section: discussion}


As shown in Section~\ref{subsection: visualization}, the more conservative it is (the larger $\beta$ is chosen), the more the reward function is underestimated. For a fixed positive $\beta$, actions that occur less often are underestimated more. When $\beta$ is sufficiently large, the conservative reward associated with $\arg\max_a \bar{\pi}(\cdot, s)$ becomes maximal. These findings corroborate our theoretical analysis and explain why CROP can alleviate the distribution shift. In fact, $\beta$ fundamentally governs the trade-off between conservatism and policy improvement. When $\beta=0$, CROP degenerates into an online RL algorithm without conservatism, and when $\beta=\infty$, CROP can be regarded as a kind of imitation learning without concern for policy improvement.
Neither extreme case provides an effective solution for offline RL problems, while an appropriately tuned intermediate $\beta$ can balance conservatism and policy improvement with satisfactory performance.
It is intuitive and widely verified that offline data with different qualities require different levels of conservatism. However, CROP and almost all existing offline RL methods still rely on heuristic-based hyperparameter tuning to achieve appropriate conservatism levels, which may hinder their applications. Designing methods with adaptive conservatism can reduce the sensitivity to hyperparameters, which will be one of the future research directions.

\begin{table}[]
\begin{center}
\caption{Computation time (unit: seconds).
}
\label{table: time}
\begin{tabular}{lll}
\hline\hline
\specialrule{0em}{2.5pt}{2.5pt}
\textbf{Dataset} & \textbf{CROP} & \textbf{RAMBO} \\ \specialrule{0em}{1.5pt}{1.5pt}
\hline
\specialrule{0em}{2.5pt}{2.5pt}
Halfcheetah-M & 116530 & 133438\\\specialrule{0em}{1.5pt}{1.5pt}
Hopper-M & 55200 & 118020\\\specialrule{0em}{1.5pt}{1.5pt}
Walker2d-M & 118680 & 132340 \\\specialrule{0em}{1.5pt}{1.5pt}
\hline
\specialrule{0em}{2.5pt}{2.5pt}
Mean & 96893 & 127933\\\specialrule{0em}{1.5pt}{1.5pt}
\hline\hline        
\end{tabular}
\end{center}
\end{table}

Although the proposed method CROP and the existing method RAMBO~\cite{RAMBO} both conservatively estimate the environment model, CROP only modifies the model training process and avoids the adversarial paradigm of updating the environment model during the policy optimization process.
Table~\ref{table: time} shows the computation time of CROP and RAMBO on three ``Medium'' datasets.
The results show that CROP takes less time to train, especially on the ``Hopper-M'' dataset.
This is because in general, the training of the environment model adopts the paradigm of supervised learning, and its gradient is more stable than that of policy optimization, requiring fewer steps. The environment of the hopper task has fewer degrees of freedom and is simpler, so the training time advantage on the ``Hopper-M'' dataset is more obvious.

Recent studies have also explored more powerful network architectures for offline RL. Notably, Q-Transformer~\cite{chebotar2023qtransformerscalableofflinereinforcement} leverages the Transformer architecture to model the Q-function through the self-attention mechanism. By integrating conservative Q-value estimation, Q-Transformer can efficiently utilize temporal information and has achieved excellent performance in large-scale robotic control benchmarks. Moreover, subsequent work~\cite{kotb2024qttdmplanningtransformerdynamics} extended the application of the Transformer to the transition model itself, enabling offline RL with end-to-end Transformer-based models. These methods demonstrate the potential of the Transformer architecture in capturing complex temporal dependencies and long-term reasoning.
Combining the development of network structure with CROP is one of the future development directions.

\section{Conclusion}
\label{section: conclusion}
This study proposes a novel model-based offline RL method, \underline{C}onservative \underline{R}eward for model-based \underline{O}ffline \underline{P}olicy optimization (CROP). The proposed method conservatively estimates the reward by concurrently minimizing both the estimation error and the rewards with random actions during model training.
Theoretical analysis shows that CROP can conservatively estimate Q-function, effectively mitigate distribution shift, and ensure a safe policy improvement. 
Experimental evaluations on D4RL benchmarks verify the performance of CROP and demonstrate its competitive results among representative offline RL approaches.
CROP provides a new perspective where online RL methods can be used on the empirical MDP with conservative rewards for offline RL problems, which is conducive to applying the latest development of online RL to offline settings.
Future work will consider hyperparameter selection without relying on online evaluation. In addition, combining model design in online model-based RL with CROP may effectively deal with more complex offline environments.


\bibliographystyle{Bibliography/IEEEtranTIE}
\typeout{}
\bibliography{Bibliography/reference}
\clearpage  
\onecolumn
\appendices
\section{Extension to Continuous State-Action Spaces}
\label{appendix:A}

{
While the theoretical analysis in Section~\ref{subsection: theoretical analysis} is primarily based on a tabular state-action space for tractability, the core mechanisms of CROP (particularly the conservative reward formulation and its impact on policy evaluation) can be naturally extended to continuous domains. This appendix outlines the theoretical rationale and key structural adaptations required for such a generalization.

First, in continuous reinforcement learning settings, the behavior policy $\bar{\pi}(a|s)$ becomes a probability density function instead of a discrete probability mass. Accordingly, the conservative reward estimator presented in Equation~\eqref{equation: reward} extends to:
\begin{equation}
r(s, a) = R(s, a) - \beta \frac{\mu}{\bar{\pi}(a|s)},
\end{equation}
where \( \mu \) denotes the density of the uniform distribution over the action space. This form continues to penalize low-density actions more severely, thereby maintaining the conservative behavior essential to mitigate the distribution shift.

Second, the Bellman operators used in our theoretical analysis can be naturally reformulated in integral form. In a continuous Markov Decision Process, the Bellman operator becomes:
\begin{equation}
(B_{T, r, \pi} Q)(s, a) = r(s, a) + \gamma \int_{\mathcal{S}} \int_{\mathcal{A}} T(s'|s, a) \pi(a'|s') Q(s', a') \, da' \, ds',
\end{equation}
which preserves the fixed-point and contraction properties necessary for policy evaluation and optimization. The results of Theorems \ref{theorem: 1} and \ref{theorem: 2} remain valid under this formulation, assuming standard regularity conditions such as boundedness and Lipschitz continuity of the reward and transition functions.

Third, the error terms involving estimation bias and sample coverage can also be defined in continuous settings using appropriate integral norms. Specifically, the quantities such as \( \epsilon_r \) and \( \epsilon_T \) can be expressed in terms of expected \( L_1 \) deviations over the empirical data distribution, while coverage assumptions are stated with respect to density functions rather than discrete counts.

Finally, although a full theoretical formalization under continuous domains (particularly when using function approximators such as neural networks) requires tools from statistical learning theory and functional analysis, our empirical results on D4RL benchmarks (which feature continuous state and action spaces) demonstrate that the conservative reward design remains effective in practice. The alignment between theoretical predictions and empirical outcomes supports the generality of our approach beyond the tabular case.

We view a rigorous extension of CROP's theoretical guarantees to continuous spaces, possibly leveraging reproducing kernel Hilbert spaces (RKHS) or generalization bounds for deep networks, as a promising direction for future work.
}

\end{document}